\documentclass[conference]{IEEEtran}

\IEEEoverridecommandlockouts % This command is only needed if you want to use the \thanks command
\usepackage[english]{babel}

\usepackage{graphicx}
\usepackage{animate}
\usepackage{epsfig} 				% for postscript graphics files
\usepackage{epstopdf}

\usepackage{listings}
\usepackage{color}
\usepackage{nameref}

\usepackage{xcolor}
\usepackage[colorlinks,bookmarks=false,citecolor=blue,linkcolor=blue,urlcolor=blue]{hyperref}
\usepackage[numbers]{natbib}	 			% change citation from [1],[2] to [1, 2]
\usepackage{amsmath}	 			% assumes amsmath package installed
\usepackage{amssymb}  				% assumes amsmath package installed

\usepackage{dsfont}			%for the symbol 'Real Numbers'
\usepackage{mathtools}

\usepackage{epigraph}
\usepackage{lscape}
\usepackage[]{nomencl}				% nomenclatures
\usepackage{algorithm}
\usepackage{algorithmic}
\usepackage{multicol}
\usepackage{multirow}
\usepackage{makecell}
\usepackage{etoolbox}
\usepackage{graphics}

\usepackage{wrapfig}

\usepackage{siunitx}

\usepackage{floatflt}

\usepackage{url}

\usepackage{placeins}
\usepackage{bm}

\usepackage{booktabs}
\usepackage{textcomp,mathcomp}
\usepackage{bm}
\usepackage{xspace}
\usepackage{mdframed}

\newcommand{\link}[1]{\colora{\url{#1}}}

\newcommand{\ProjectWeb}[0]{\href{https://linchangyi1.github.io/LightTact
}{https://linchangyi1.github.io/LightTact}}

\newcommand{\sensor}[0]{LightTact\xspace}

\begin{document}

% paper title
\title{\LARGE \bf \sensor: A Visual–Tactile Fingertip Sensor for \\ Deformation-Independent Contact Sensing}

% authors
\author{
Changyi Lin$^{*1}$, Boda Huo$^{*1}$, Mingyang Yu$^{1}$, Emily Ruppel$^{2}$,
Bingqing Chen$^{2}$, Jonathan Francis$^{1,2}$,  Ding Zhao$^{1}$
\vspace{0.1cm}
\\
\ProjectWeb
\thanks{$^{*}$ Equal contribution}
\thanks{$^{1}$ Carnegie Mellon University}
\thanks{$^{2}$ Bosch Center for Artificial Intelligence (BCAI)}
}

\maketitle
\IEEEpeerreviewmaketitle

%===============================================================================
\begin{abstract}
Contact often occurs without macroscopic surface deformation, such as during interaction with liquids, semi-liquids, or ultra-soft materials.
However, most existing tactile sensors rely on deformation to infer contact, making such light-contact interactions difficult to perceive robustly.
To address this, we present \sensor, a visual-tactile fingertip sensor that makes \emph{contact directly visible} via a deformation-independent principle.
\sensor features an ambient-blocking optical configuration that suppresses both external light and internal illumination at non-contact regions, while transmitting only the scattered light generated at true contacts.
As a result, \sensor produces high-contrast raw images in which non-contact pixels remain near-black (mean gray value $<3$) and contact pixels preserve the natural appearance of the contacting surface.
Built on this, \sensor achieves accurate pixel-level contact segmentation that is robust to material properties, contact force, surface appearance, and environmental lighting.
We further demonstrate that \sensor unlocks new robotic manipulation behaviors that require detection of extremely light contact, including water spreading, facial-cream dipping, and soft thin-film interaction.
In addition, we show that \sensor's spatially aligned visual-tactile images can be directly interpreted by vision-language models.
\end{abstract}

\section{Introduction}
\label{sec:intro}
Tactile sensing is essential for robotic manipulation, where robots must precisely regulate when, where, and how they physically interact with objects through contact~\cite{dahiya2009tactile, luo2025tactile}.
In recent years, vision-based tactile sensors (VBTSs) have gained prominence due to their high spatial resolution and low cost~\cite{li2025classification, li2024vision, xin2025vision, he2025survey}.
However, most VBTSs sense contact \emph{indirectly} by observing macroscopic deformation of a soft surface~\cite{yuan2017gelsight, yamaguchi2016combining, lin20239dtact, ward2018tactip, zhang2022deltact}.
This deformation-dependent mechanism fundamentally requires measurable indentation to generate an informative tactile signal.
In contrast, many real-world interactions involve \emph{light} contacts that induce little or no surface deformation.
Examples include interactions with liquids or semi-liquids, and gentle contacts with extremely soft or thin materials.
Such scenarios are common in everyday tasks, such as spreading water on a tabletop, dipping facial cream, or detecting a sheet of food film, but they remain challenging for deformation-based tactile sensors.

To relax this constraint, several sensors attempt to bypass deformation by using frustrated total internal reflection (TIR) to highlight the contact regions of a transparent medium~\cite{han2005low, shimonomura2016robotic, zhang2023tirgel, sun2025soft}.
However, ambient illumination and reflections from non-contact regions can also enter the camera, seriously corrupting the highlighted contact signal.
As a result, TIR-based sensors typically function reliably only with monochromatic objects or under dark, tightly controlled environments---conditions incompatible with robotic tactile perception in everyday, unstructured settings.
Therefore, designing a tactile sensor that can robustly detect contact \emph{without} relying on surface deformation remains an open challenge.

\begin{figure}[t]
\centering
\includegraphics[width=\linewidth]{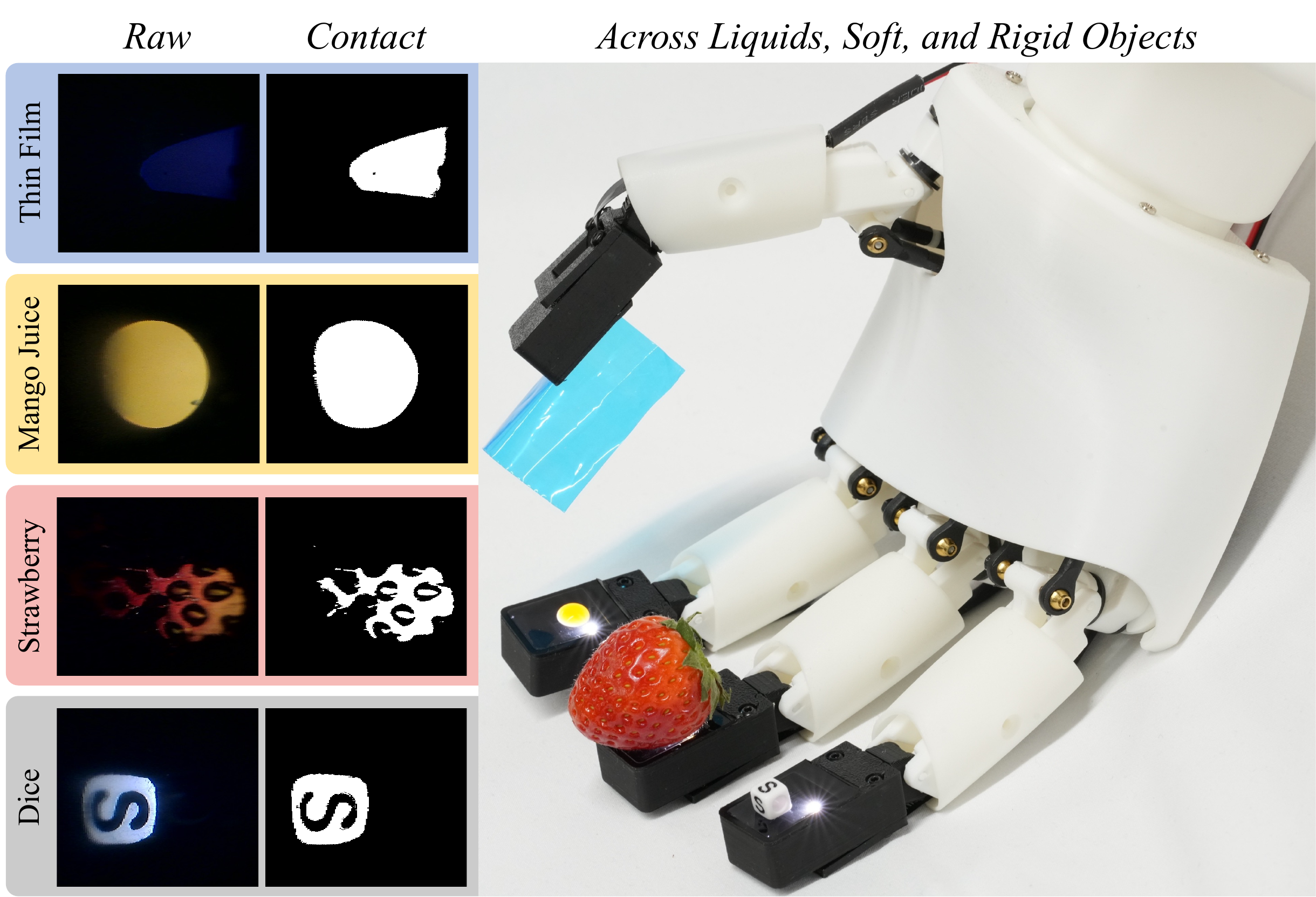}
\vspace{-0.6cm}
\caption{\sensor provides direct, pixel-level contact sensing across liquids, ultra-soft materials, and rigid objects, without requiring a minimum contact force.
Its optical design produces high-contrast raw images in which non-contact pixels remain near-black, while contact pixels preserve the natural appearance of the contacting surface.
\sensor is fingertip-sized and can be integrated into dexterous hands such as Amazing Hand~\cite{pollenrobotics_amazinghand_github_2025}.
}
\vspace{-0.4cm}
\label{fig:teaser}
\end{figure}

In this work, we introduce \sensor, a camera-based fingertip sensor that makes physical contact \emph{directly visible} in the raw image.
\sensor enforces a bijective contact-visibility relationship: a pixel is visible \emph{if and only if} it is in true physical contact.
This enables robust pixel-level contact segmentation across liquids, ultra-soft materials, and rigid objects under diverse object appearances and bright ambient lighting (Fig.~\ref{fig:teaser}).

The key enabler is our \textit{ambient-blocking} optical configuration that can, in theory, completely suppress both external light and internal illumination at non-contact regions, while transmitting only contact-generated scattered light to the camera through a transparent medium.
We realize this behavior by adopting a side-view imaging optical layout with dedicated camera-LED-medium configurations, which jointly leverage refraction, specular reflection, and total internal reflection (TIR) to block non-contact light paths.
To translate this principle into a practical robot fingertip, we co-design \sensor across optics, structure, materials, electronics, and fabrication.
Finally, \sensor features the following properties:
\begin{itemize}
    \item \textbf{High-contrast raw images} where non-contact pixels remain near-black (gray value~$<3$) while contact pixels preserve the surface appearance.
    \item \textbf{Pixel-level contact segmentation} with lightweight computation, robust to material properties, contact force, ambient lighting, and surface appearance.
    \item \textbf{Aligned visual–tactile signals} where both local appearance of the object and contact region of the sensor surface are naturally co-registered in the same RGB image.
    \item \textbf{Compact and compliant.} \sensor has a fingertip-scale size of $12 \times 18 \times 34.5~\text{mm}$ and a soft surface, enabling direct integration into grippers and dexterous hands.
    \item \textbf{Affordable and reproducible.} The primary cost is the RGB camera; the remaining components can be fabricated with standard tools for under $\$20$.
    \item \textbf{Open-source.} We open-source both the hardware and software, together with a step-by-step fabrication tutorial to facilitate reproduction and further development.
\end{itemize}

Leveraging these properties, we integrate \sensor into a robotic system and demonstrate manipulation behaviors that remain difficult or infeasible for current advanced tactile sensors.
In particular, \sensor enables reliable perception of extremely light contact during interaction with liquids, semi-liquids, and ultra-soft or thin materials, where deformation-based VBTSs and force sensing provide uninformed or weak signals.
Moreover, we demonstrate that \sensor's visual–tactile outputs can be directly interpreted by Vision-Language Models (VLMs), enabling fine-grained reasoning such as resistor value inference for robotic sorting.

\section{Related Work}
\label{sec:related}
We review prior work in two aspects most relevant to this paper:
(1) \textit{robust contact sensing and segmentation} in vision-based tactile sensors (VBTSs), and
(2) \textit{simultaneous visual–tactile sensing} for robotic manipulation.

\subsection{Contact Sensing with Vision-Based Tactile Sensors}
Vision-based tactile sensors (VBTSs) use internal cameras to image tactile signals with high spatial sensitivity at low cost, but most detect contact \emph{indirectly} via macroscopic deformation of a compliant surface.
Common approaches include tracking painted markers~\cite{yamaguchi2016combining, ward2018tactip, alspach2019soft, cui2021hand, zhang2022deltact, zhang2022tac3d}, observing magnified reflective patterns~\cite{saga2007high, saga2013precise}, or measuring luminance variations of reflective layers~\cite{yuan2017gelsight, lin20239dtact}.
Because they require visible deformation, the surface must be soft, the contacting object relatively harder, and the applied force above a detectable threshold, limiting \emph{light contacts} with negligible indentation (e.g., water, facial cream, thin films).

To decouple sensing from deformation, several sensors~\cite{han2005low, shimonomura2016robotic, zhang2023tirgel, sun2025soft} apply total internal reflection (TIR) within a transparent medium; contact locally frustrates TIR and increases reflected light, highlighting contact regions.
However, this strategy is sensitive to ambient illumination: external light, reflections from non-contact portions of the object, and internal LED leakage can enter the camera through non-contact regions, corrupting the intended TIR signal.
As a result, these systems often require monochromatic objects or controlled dark environments.

In contrast, \sensor uses an ambient-blocking optical layout that suppresses both external and internal illumination at non-contact regions while capturing only contact-generated scattered light, keeping non-contact pixels near-black in raw images and enabling robust pixel-level contact segmentation across lighting conditions and object appearances.

Similar objectives appear in non-robotic optical devices, such as the Wyman--White fingerprint imager~\cite{wyman1965method} and the Drawing Prism~\cite{greene1985drawing}.
However, these systems target static capture or tracing and assume more relaxed constraints (e.g., controlled lighting and limited object types), often relying on bulky geometries, dark environments, or even a thin liquid layer to remove air gaps---conditions incompatible with dynamic robotic tactile perception in everyday settings. These conditions are incompatible with robotic tactile perception.

\subsection{Visual-Tactile Multimodal Sensing}
A prevalent strategy for combining visual and tactile modalities is to paint markers on a transparent gel~\cite{yamaguchi2016combining, hogan2022finger, wang2022spectac, fan2024vitactip, luo2024compdvision, nguyen2025vi2tap, li2025simultaneous}. However, this approach provides only sparse motion cues and inherently occludes visual information at marker locations.
To preserve full-pixel visibility, alternative methods utilize translucent coatings paired with illumination switching~\cite{hogan2021seeing, roberge2023stereotac, athar2023vistac}, but they support only one modality at a time.
While TIR-based designs~\cite{han2005low, shimonomura2016robotic, zhang2023tirgel, sun2025soft} can theoretically provide visual information, the bright internal illumination required for tactile sensing often causes overexposure, degrading the external visual features.
In contrast, \sensor achieves the simultaneous acquisition of robust pixel-level contact segmentation and natural visual appearance using a single camera, ensuring both modalities are spatially and temporally aligned by construction.

\section{Methodology}
\label{sec:methodology}
Our goal is to design \sensor as a robotic tactile sensor capable of reliable, pixel-level contact segmentation across diverse objects and lighting conditions.
To this end, we first introduce a novel optical layout and sensing principle and analyze its theoretical behavior in Section~\ref{subsec:sensing_principle}.
We then present how \sensor realizes this principle within a compliant, fingertip-scale form factor suitable for robotic integration in Section~\ref{subsec:sensor_design}.
Following, we describe the key fabrication steps for constructing \sensor sensors in Section~\ref{subsec:sensor_fabrication}.
Finally, we introduce the procedures for sensor calibration and algorithms for contact segmentation in Section~\ref{subsec:contact_segmentation}.

% \newpage
\subsection{Optical Layout and Sensing Principle}
\label{subsec:sensing_principle}

\begin{figure}[t]
\centering
\includegraphics[width=\linewidth]{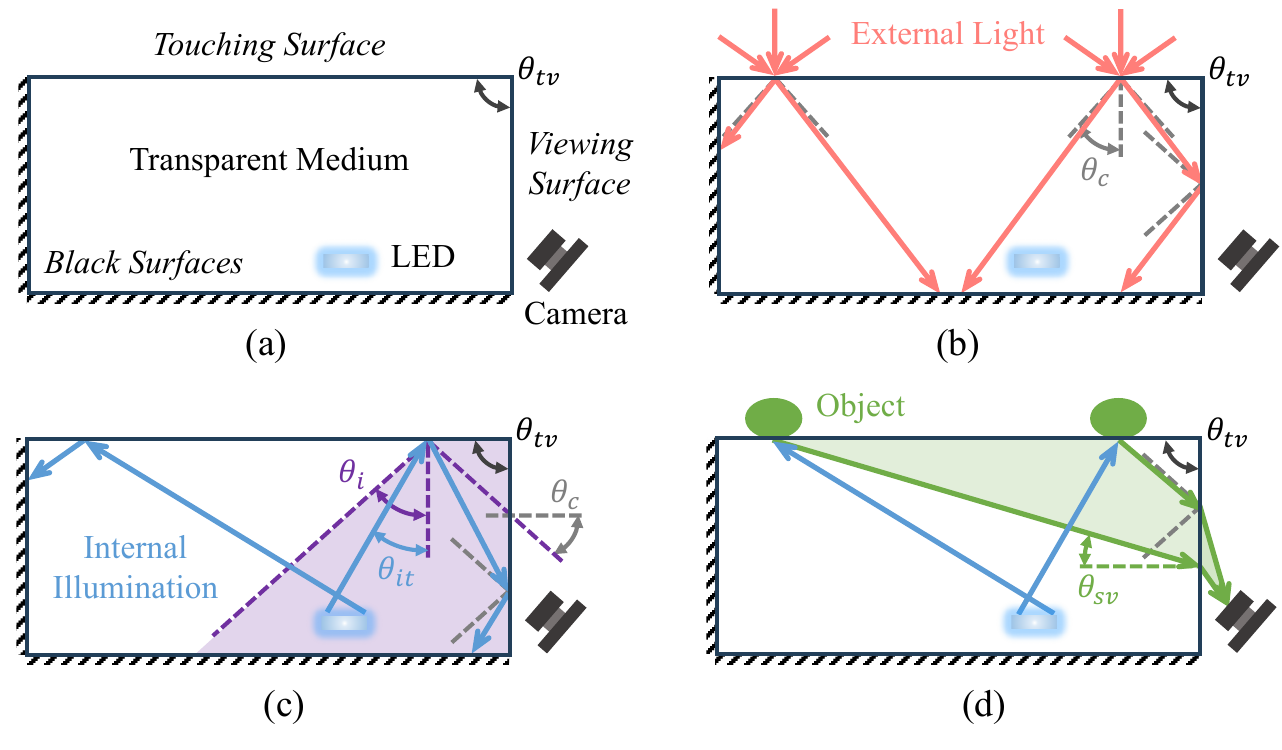}
\vspace{-0.8cm}
\caption{Optical layout and sensing principle of \sensor.
(a) Core components: a transparent medium with a touching and viewing surface, an internal LED, and a camera.
(b) External light entering through non-contact regions is rejected by refraction followed by total internal reflection (TIR).
(c) Internal LED illumination reflecting from non-contact regions is also rejected by TIR when the LED is placed within the purple area to satisfy $\theta_{it}<\theta_i$.
(d) At true contact, the air gap is removed and the contacting surface produces diffuse scattering. A subset of these rays reaches the viewing surface with incidence $<\theta_c$ and refracts toward the camera.}
\vspace{-0.4cm}
\label{fig:sensing_principle}
\end{figure}

Figure~\ref{fig:sensing_principle} (a) shows the core optical elements of \sensor:  
a transparent medium that interacts with external objects, an internal LED for illumination, and a camera that observes light exiting the medium.  
Two surfaces of the medium play distinct optical roles: the \emph{touching surface}, where contact occurs, and the \emph{viewing surface}, through which light exits toward the camera.
The remaining surfaces are coated matte-black to absorb any light that reaches them.
At the touching surface, combining two light sources (external light and internal) with two region types (contact and non-contact) produces three relevant light behaviors for non-luminous objects:
(1) external light incident on non-contact regions,
(2) internal illumination reflected on non-contact regions, and
(3) internal illumination scattered at contact regions.

Our objective is to obtain deformation-independent contact segmentation from the raw image.
To achieve this, the optical layout must \emph{completely suppress} external and internal light at non-contact regions, while \emph{transmitting} only light generated at true contacts.
The key is a side-view imaging layout: the camera observes the medium through a viewing surface that forms an angle $\theta_{tv}$ with the touching surface, rather than being parallel.
This non-parallel geometry separates the light paths, so that only contact-generated light can transmit through the viewing surface and reach the camera.
Table~\ref{tab:sensing_principle_table} summarizes these behaviors, and we detail them below.

\paragraph{External light rejection at non-contact regions} 
As illustrated in Fig.~\ref{fig:sensing_principle} (b), external rays approach the touching surface from a wide range of directions. The sensing region corresponds to the span between the two depicted ray bundles.
When external light enters the medium through non-contact areas, it refracts at angles below the critical angle $\theta_c$ due to the air–medium index difference.
A portion of these refracted rays then propagate toward the viewing surface.
By choosing a sufficiently large wedge angle ($\theta_{tv}>2\theta_c$), all such rays arrive at the viewing surface with incidence $>\theta_c$, causing them to undergo total internal reflection (TIR) and redirecting them toward the black surfaces.
Consequently, no external light from non-contact regions reaches the camera.

\paragraph{Internal illumination rejection at non-contact regions}  
Internal LED illumination should contribute to the image only at true contacts.
As shown in Fig.~\ref{fig:sensing_principle} (c), LED rays that reflect specularly off non-contact regions of the touching surface, must also reach the viewing surface with incidence $>\theta_c$ to ensure TIR.  
This is guaranteed by constraining the LED position: the LED must lie entirely within the purple admissible region such that all emitted rays satisfy $\theta_{it} < \theta_i = \theta_{tv}-\theta_c$.  
Under this placement rule, every specularly reflected LED ray is rejected by TIR at the viewing surface.

\begin{table}[t]
\centering
\setlength{\tabcolsep}{2pt}
\caption{Optical behaviors and geometric conditions for each light–surface interaction.}
\begin{tabular}{ccccc}
\toprule
\multirow{2}{*}{Source}
& \multirow{2}{*}{Region}
& \multicolumn{2}{c}{Optical behavior at surfaces} 
& \multirow{2}{*}{Design condition} \\
\cmidrule(lr){3-4}
 & & Touching surface & Viewing surface &  \\ 
\midrule
External & Non-con & Refraction & TIR & $\theta_{tv} > 2\theta_c$ \\
Internal & Non-con & Specular reflection & TIR & $\forall \theta_{it} < \theta_{tv}-\theta_c$ \\
Internal & Contact & Diffuse scattering & Refraction & $\theta_{tv} < \frac{\pi}{2}+\theta_c$ \\
\bottomrule
\end{tabular}
% \vspace{-0.4cm}
\label{tab:sensing_principle_table}
\end{table}

\paragraph{Appearance capture at contact regions}
When an object touches the medium, the air gap disappears and the contacting surface produces diffuse scattering of the LED illumination (Fig.~\ref{fig:sensing_principle}(d)).
This generates a wide angular spread of rays, which can reach the viewing surface with incidence $\theta_{sv} \in \left(\theta_{tv}-\frac{\pi}{2}, \frac{\pi}{2} \right)$.
To observe the contacting surface, a subset of these scattered rays must refract through the viewing surface and reach the camera.
This requires that the incidence range $\left(\theta_{tv}-\frac{\pi}{2}, \frac{\pi}{2} \right)$ intersects the transmissible range $\left(-\theta_{c}, \theta_{c} \right)$, which is satisfied when $\theta_{tv}<\frac{\pi}{2}+ \theta_c$.
Under this condition, a subset of diffuse-scattering rays transmits within the green regions and enters the camera, revealing the natural local appearance of the object.

Our optical layout enforces $\theta_{tv} > 2\theta_c$ so that any light entering through non-contact regions arrives at the viewing surface with incidence $>\theta_c$ and is therefore rejected by TIR, preventing it from reaching the camera.
Strictly speaking, $\theta_{tv}>2\theta_c$ is not the only feasible configuration: when $\theta_{tv}<2\theta_c$, the camera could, in principle, be placed outside the refracted ray field from non-contact regions.
In practice, however, this alternative sacrifices compactness and robustness, as it requires a larger camera offset and stricter suppression of stray reflections.
We therefore adopt $\theta_{tv}>2\theta_c$ as a simple and reliable default; a detailed analysis of the $\theta_{tv}<2\theta_c$ case is provided in Appendix~\ref{app:camera_placement}.

\subsection{Sensor Design}
\label{subsec:sensor_design}

\begin{figure}[ht]
\centering
\includegraphics[width=\linewidth]{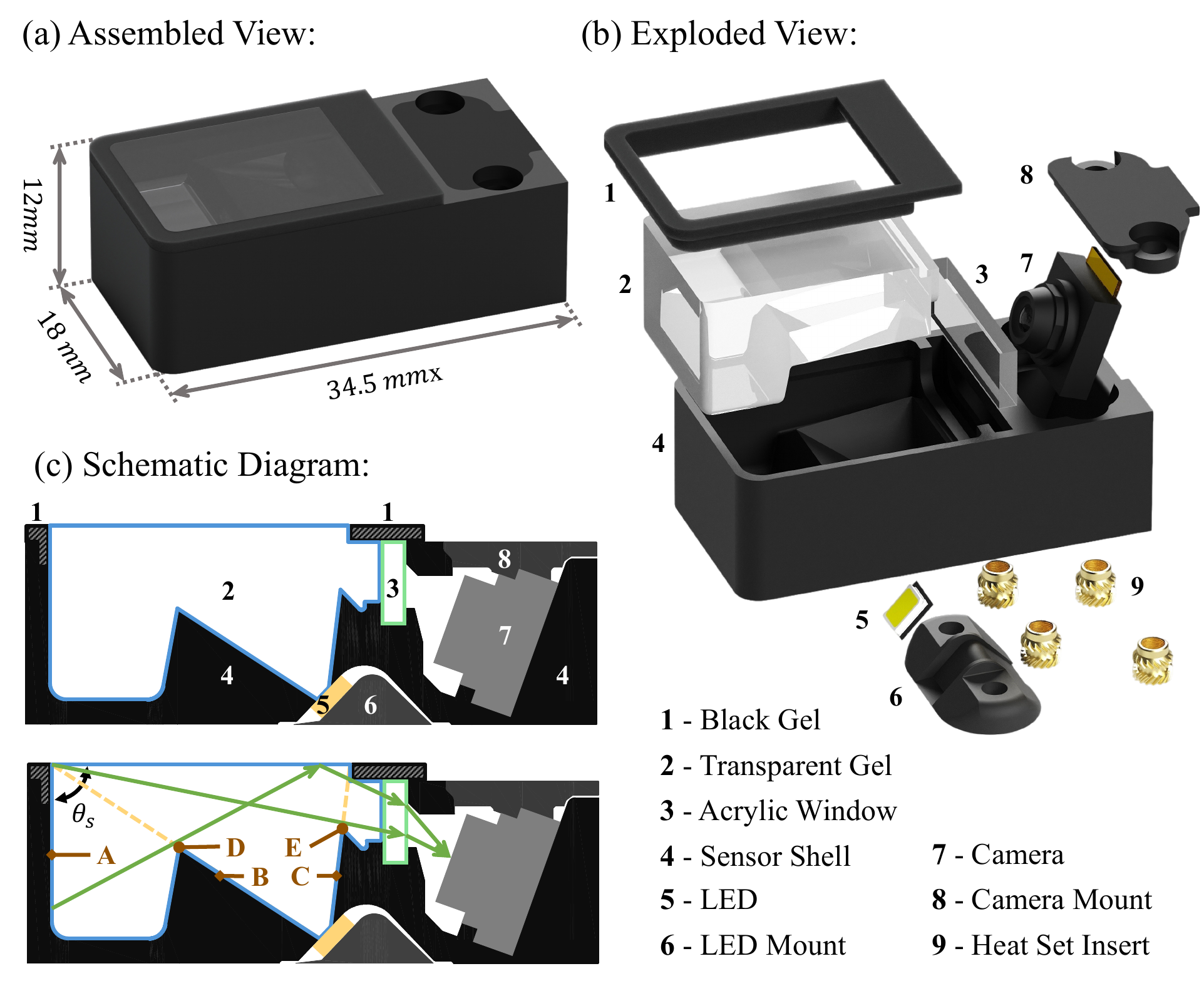}
\vspace{-0.6cm}
\caption{Design of \sensor.
(a) Assembled view.
(b) Components of \sensor illustrated in exploded view.
(c) Schematic diagram.
}
\vspace{-0.4cm}
\label{fig:sensor_design}
\end{figure}

As shown in Fig.~\ref{fig:sensor_design}(a), \sensor is extremely compact, with a fingertip-scale size of only $12 \text{mm} \times 18 \text{mm}\times 34.5 \text{mm}$.
Figures~\ref{fig:sensor_design}(b) and~\ref{fig:sensor_design}(c) illustrate the internal assembly, which mirrors the optical system described in Section~\ref{subsec:sensing_principle}:
a transparent medium (2, 3), an internal LED (5), and a camera (7), all housed within a rigid shell (4), a surrounding black gel layer (1), and lightweight mounts (6, 8).
The following paragraphs describe each component in detail.

\textbf{Transparent gel and acrylic window.}
For most robotic applications, the touching surface should be compliant to support safe and stable interaction.
Importantly, this compliance is a design preference rather than a sensing requirement, since our optical principle does not rely on macroscopic deformation.
However, if the entire transparent medium is soft, the viewing surface would deform under load and introduce image distortion.
To avoid this, we construct the medium as a composite: a soft, transparent gel that forms the touching surface, and a rigid acrylic window that defines a deformation-free viewing surface.
The two materials have similar refractive indices ($\approx 1.45$), corresponding to a critical angle of $\theta_c \approx 43^\circ$.
To satisfy the geometric conditions in Section~\ref{subsec:sensing_principle}, namely $\theta_{tv}>2\theta_c\approx86^\circ$ to suppress non-contact light and $\theta_{tv}<90^\circ+\theta_c\approx133^\circ$ to transmit contact-scattered light, we set the touching and viewing surfaces to be perpendicular.

\textbf{LED and camera.}
Internal illumination is provided by a single 2835-SMD LED, mounted at an oblique angle ($\approx45^\circ$) to improve lighting uniformity across the touching surface.
Imaging is performed by a compact UVC camera (OV5693) with a $120^\circ$ field of view, placed close to the transparent medium to preserve a fingertip-scale form factor.
Both components are secured with lightweight mounts that are attached to the sensor shell using M2 screws and heat-set inserts.

\textbf{Sensor shell.}
The shell not only holds the optical components at their required poses, but also enforces three key geometric-optical constraints: the transparent gel's geometry, the LED's illumination envelope, and the camera's effective viewing range. These design aspects are detailed below.

\paragraph{Transparent gel geometry}
Although the shell is printed in black to approximate the ideal light-absorbing boundaries, practical materials still exhibit residual reflectance.
As shown in Fig.~\ref{fig:sensor_design}(c), surface~A can reflect stray light into the camera through non-contact regions.
To mitigate this, we design surface~A to be vertical ($\theta_s=\frac{\pi}{2}$), reducing the chance that specular reflections couple into the viewing path.
For applications that require access to confined spaces, the shell can be made wedge-shaped.
We analyze the case's ($\theta_s<\frac{\pi}{2}$) optical consequences in Appendix~\ref{app:wedge_shape}.

\paragraph{LED illumination range}
To prevent the LED from directly illuminating surface~A, the shell includes two baffles (surfaces~B and~C) that bound the emission cone.
Increasing baffle height improves shielding but is limited by two factors:
(1) if surface~B extends above point~D, the exposed inclined region becomes directly illuminated and increases non-contact light leakage; and
(2) if surface~C exceeds point~E, it occludes part of the contact-generated scattered light from the left portion of the touching surface.

\paragraph{Camera’s effective viewing range}
The shell also apertures the camera so that it observes only the touching surface.
Allowing the camera to view outside this region would admit additional stray light (e.g., scattering from the rough top or bottom faces of the acrylic window) and interfere the pixels associated with the touching surface.

\textbf{Black gel.}
To maintain compliance across the full contact area, we cast a soft black gel around the perimeter of the transparent gel, creating a smooth transition between the touching surface and the rigid shell.
It also covers the top surface of the acrylic window, which is too rigid and rough to serve as part of the touching surface.

\subsection{Sensor Fabrication}
\label{subsec:sensor_fabrication}

\begin{figure}[t]
\centering
\includegraphics[width=\linewidth]{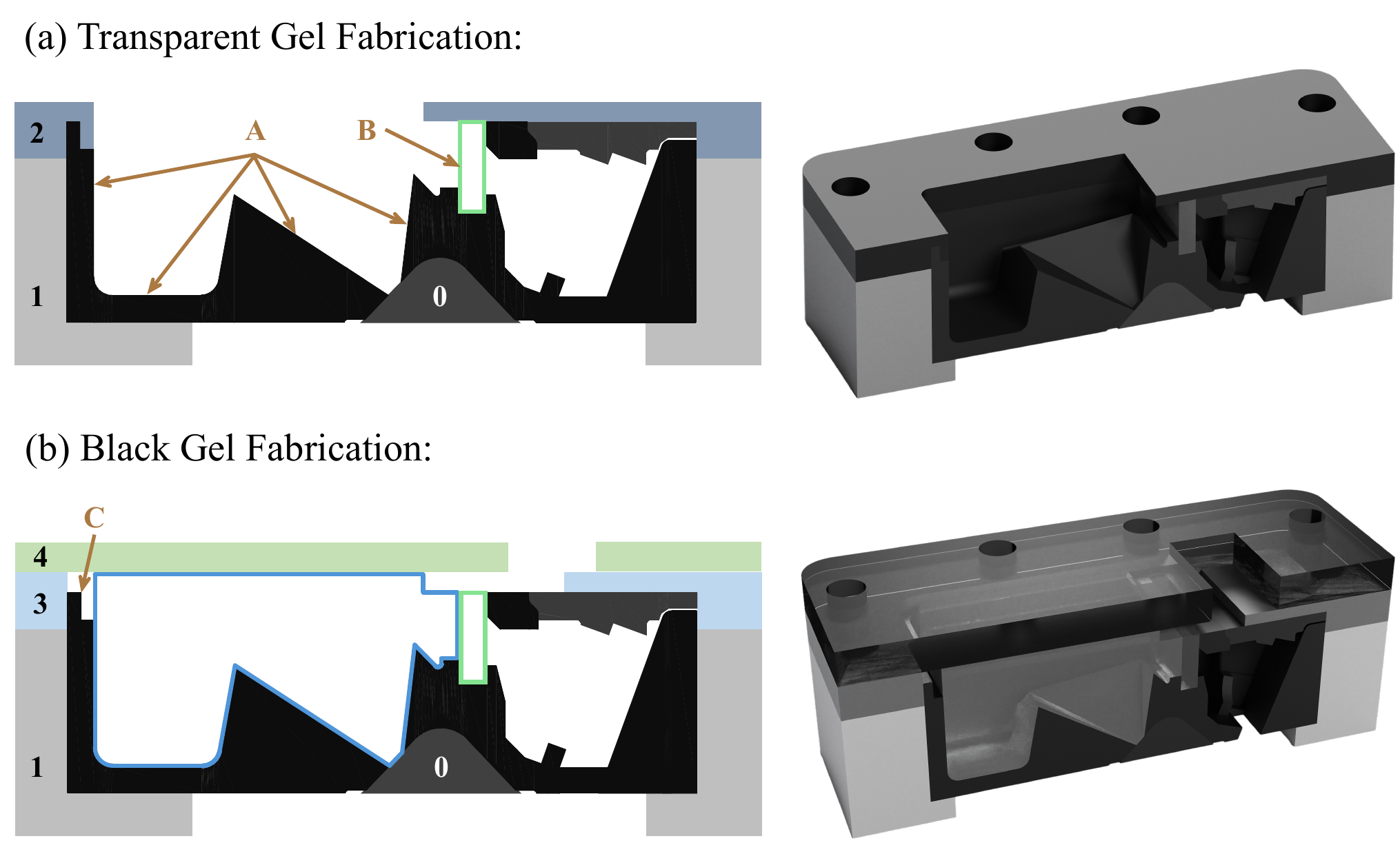}
\caption{Gel fabrication for \sensor.
(a) Casting of the transparent gel. The sensor shell, the mold~0, and the acrylic window together form a closed container.
(b) Casting of the black gel. The top acrylic mold~4 ensures that the black gel cures to a uniform height around the transparent gel.
}
\vspace{-0.4cm}
\label{fig:sensor_fabrication}
\end{figure}
Most components of \sensor are straightforward to fabricate: the shell and mounts are 3D-printed with black PLA, the acrylic window is laser-cut from $1.5~\text{mm}$ acrylic sheet, the LED board is made by FPCB services, and the camera is off-the-shelf.
% In our build, the components excluding the camera cost under $\$20$ in total.
The only nontrivial steps are casting the two gel layers, which require auxiliary molds and temporary inserts to define accurate internal boundaries (Fig.~\ref{fig:sensor_fabrication}).

\textbf{Transparent gel.}
We choose Silicones Inc\textsuperscript{\textregistered} XP-565 silicone (cured hardness Shore~A27) to make the transparent gel.
The acrylic window is installed in the shell to form a sealed container for casting.
To prevent silicone leakage into the LED compartment, we temporarily fill the LED recess with a solid insert (mold~0 in Fig.~\ref{fig:sensor_fabrication}(a)).
Two additional 3D-printed molds (labeled~1 and~2) are stacked around the shell to reserve the perimeter volume occupied by the future black gel.

Before pouring, we brush a high-absorption black pigment onto the relevant inner shell surfaces (region~A in Fig.~\ref{fig:sensor_fabrication}(a)) to better approximate the light-absorbing boundary condition.
To ensure strong bonding at the gel-acrylic interface, we apply a thin layer of silicone adhesive (Smooth-On\textsuperscript{\textregistered} Sil-Poxy) to the acrylic face labeled~B.
We then degas and pour the XP-565 mixture until the cavity is fully filled, and cure it for $\sim$4~hours on a heated 3D-printer bed at $50^\circ\mathrm{C}$.

\textbf{Black gel.}
The black gel is made from the same XP-565 silicone mixed with black pigment.
After the transparent gel cures, we remove mold~2 and replace it with a second top mold (mold~3 in Fig.~\ref{fig:sensor_fabrication}(b)) that defines the perimeter cavity for the black gel.
Because this cavity is narrow and difficult to fill uniformly from above, we place an additional acrylic mold (labeled~4) on top of mold~3 and pour the mixture through the opening in mold~4, allowing it to flow uniformly around the perimeter.
This ``cap'' mold also keeps firmly pressed against the touching surface, ensuring a consistent cured height.
After curing under the same heated-bed conditions, we remove the molds and trim any excess gel around the fill opening.

As indicated by label~C in Fig.~\ref{fig:sensor_fabrication}(b), the shell includes a small raised perimeter step that creates the final three-layer edge profile (“shell – black gel – transparent gel”).
This step is necessary because mold~4 slightly compresses the transparent gel during black-gel casting; after demolding, the transparent gel elastically recovers and can leave a thin lateral air gap beneath the black gel.
Without the raised step, this gap would form a direct light path into the sensor.
The raised perimeter therefore acts as a light-blocking barrier, preserving the dark-boundary assumption required by our optical design.

\newcommand{\widthenv}{1.29cm}
\newcommand{\rownumenv}{3}
\newcommand{\lengthref}{1.25cm}
\newcommand{\rownumber}{4}
\begin{table*}[ht]
\centering
\setlength{\tabcolsep}{2pt}
\caption{Intensity statistics and representative images of non-contact and contact conditions under different internal and external illumination levels.}
\begin{tabular}{lcccccccccc}
\toprule
Internal brightness (Lux) & 0 & 430 & 1050 & 1670 & 430 & 430 & 430 & 430 & 430 & 430 
\\
External brightness (Lux) & 3 & 3 & 3 & 3 & 740 & 1020 & 1510 & 2010 & 2540 & 3520 
\\
\multicolumn{1}{l}{\multirow{\rownumenv}{*}{\sensor in environment}}
& \multirow{\rownumenv}{*}{\includegraphics[width = \widthenv]{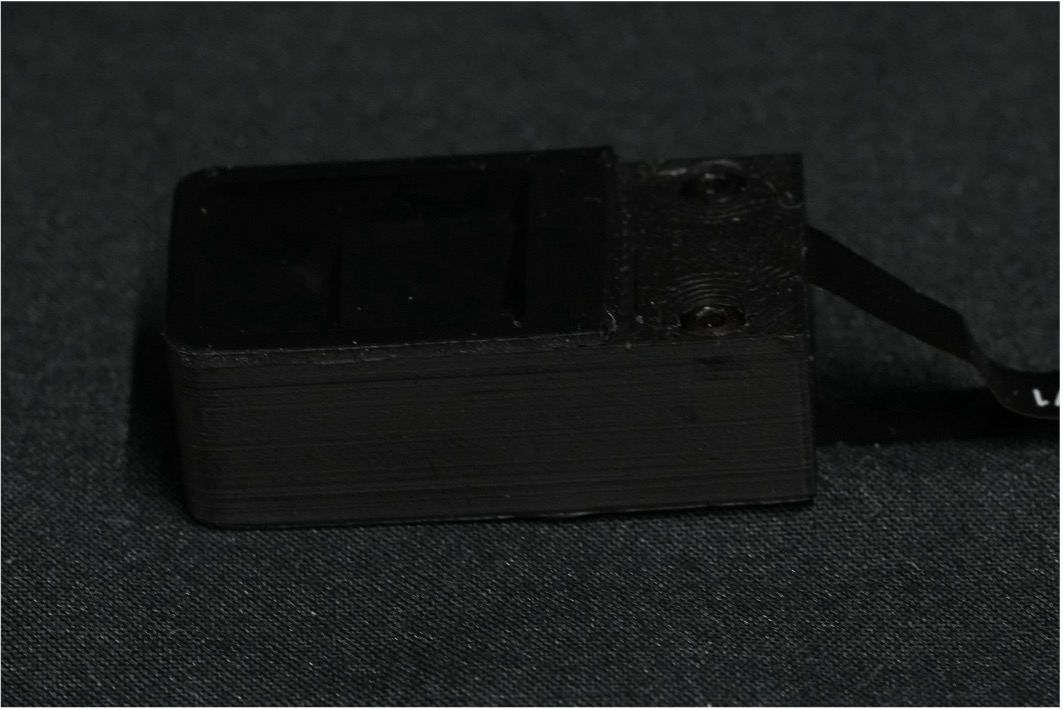}}
& \multirow{\rownumenv}{*}{\includegraphics[width = \widthenv]{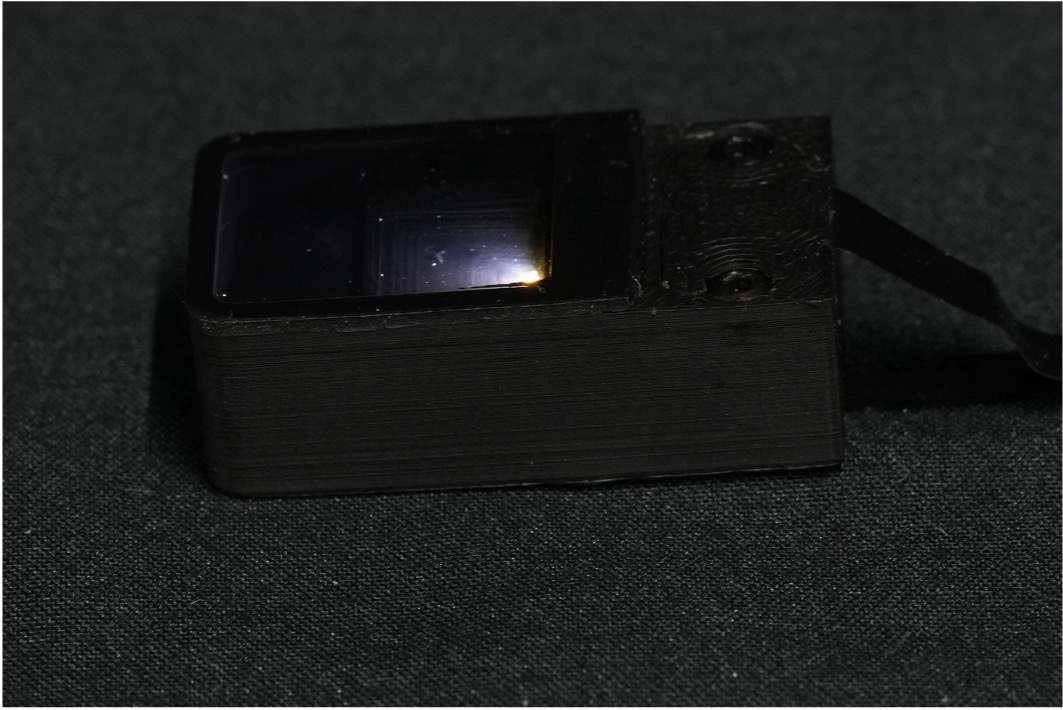}}
& \multirow{\rownumenv}{*}{\includegraphics[width = \widthenv]{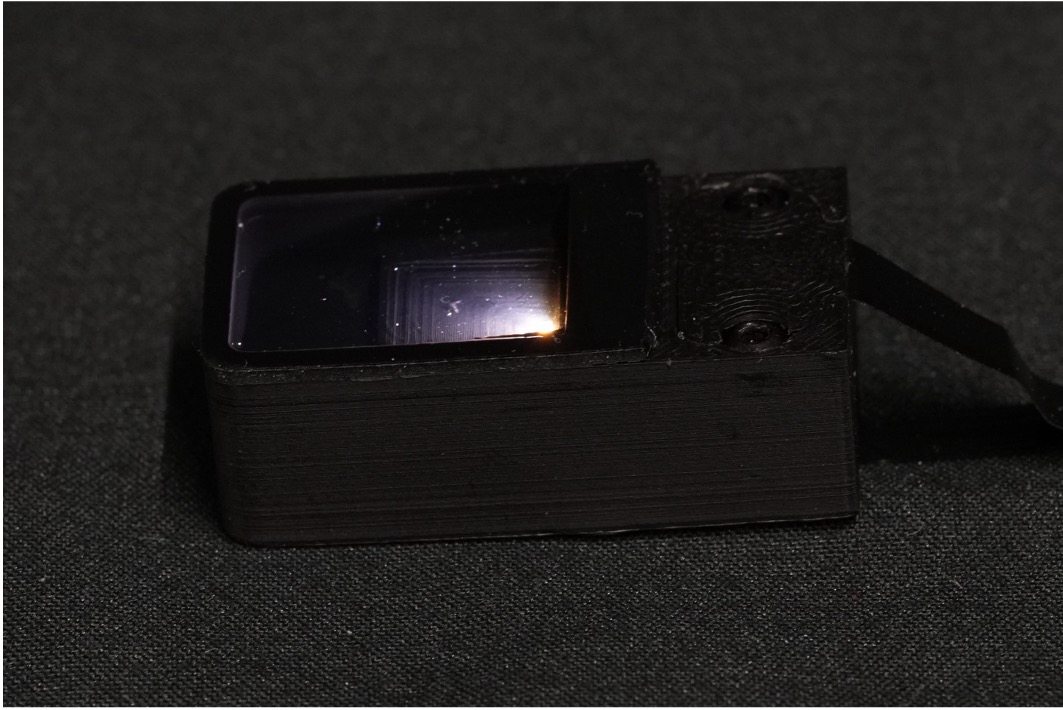}}
& \multirow{\rownumenv}{*}{\includegraphics[width = \widthenv]{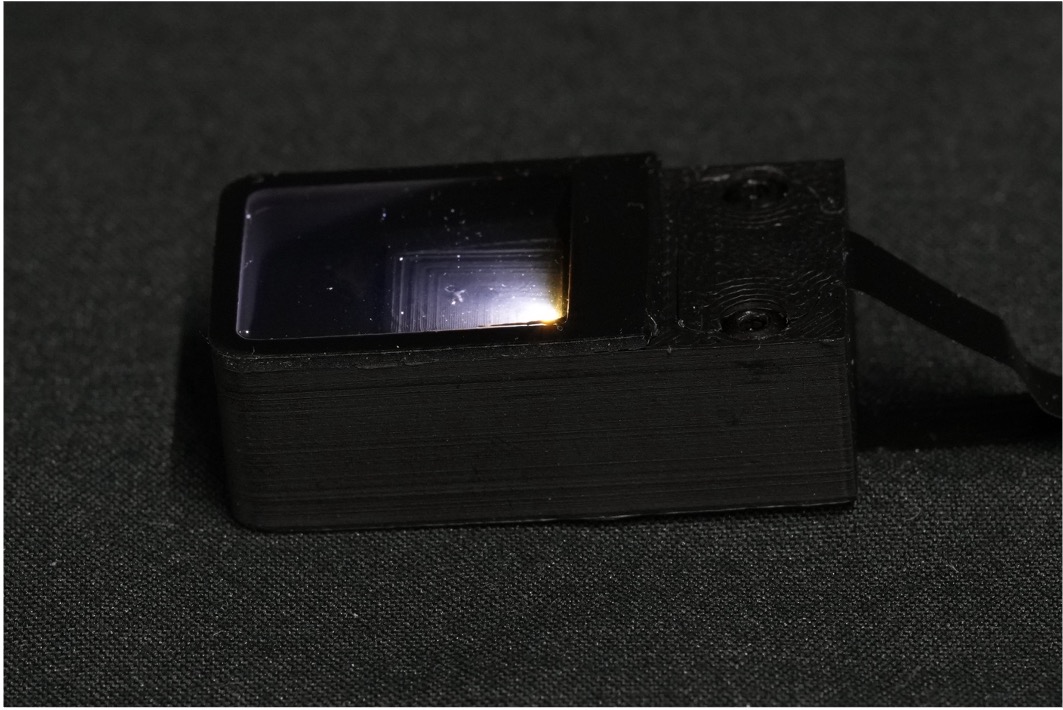}}
& \multirow{\rownumenv}{*}{\includegraphics[width = \widthenv]{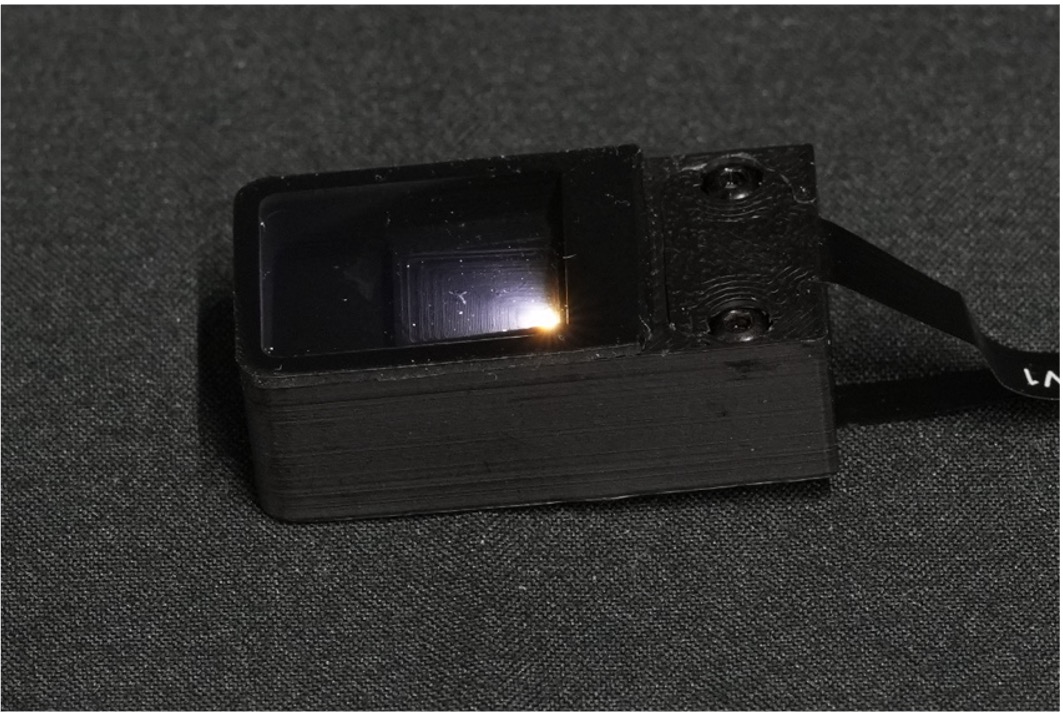}}
& \multirow{\rownumenv}{*}{\includegraphics[width = \widthenv]{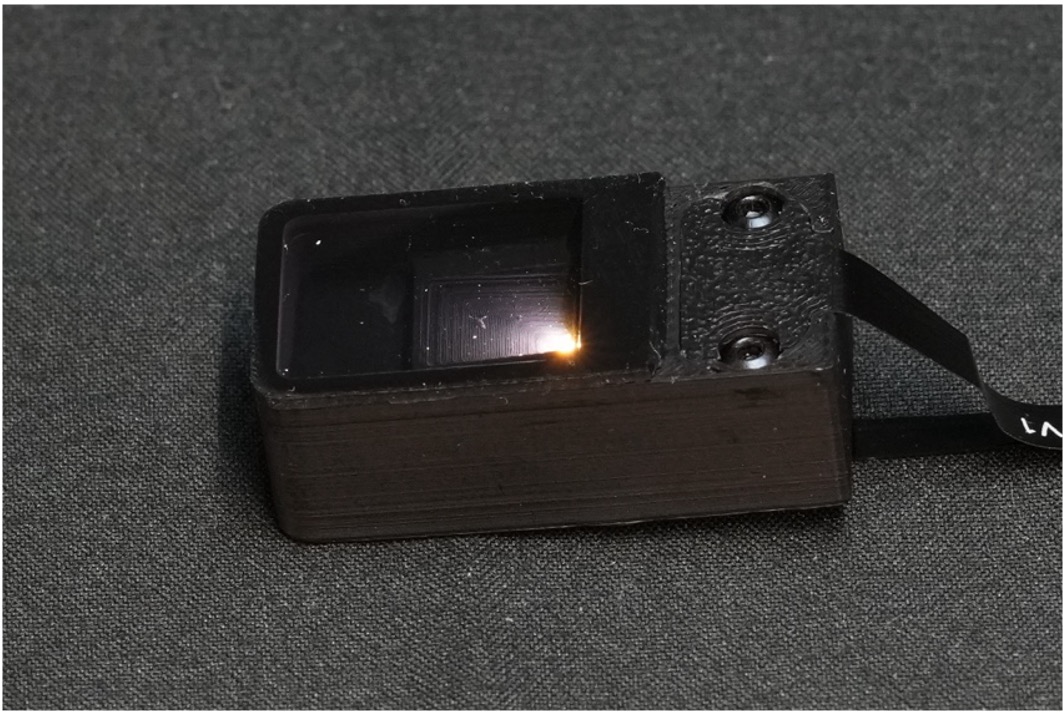}}
& \multirow{\rownumenv}{*}{\includegraphics[width = \widthenv]{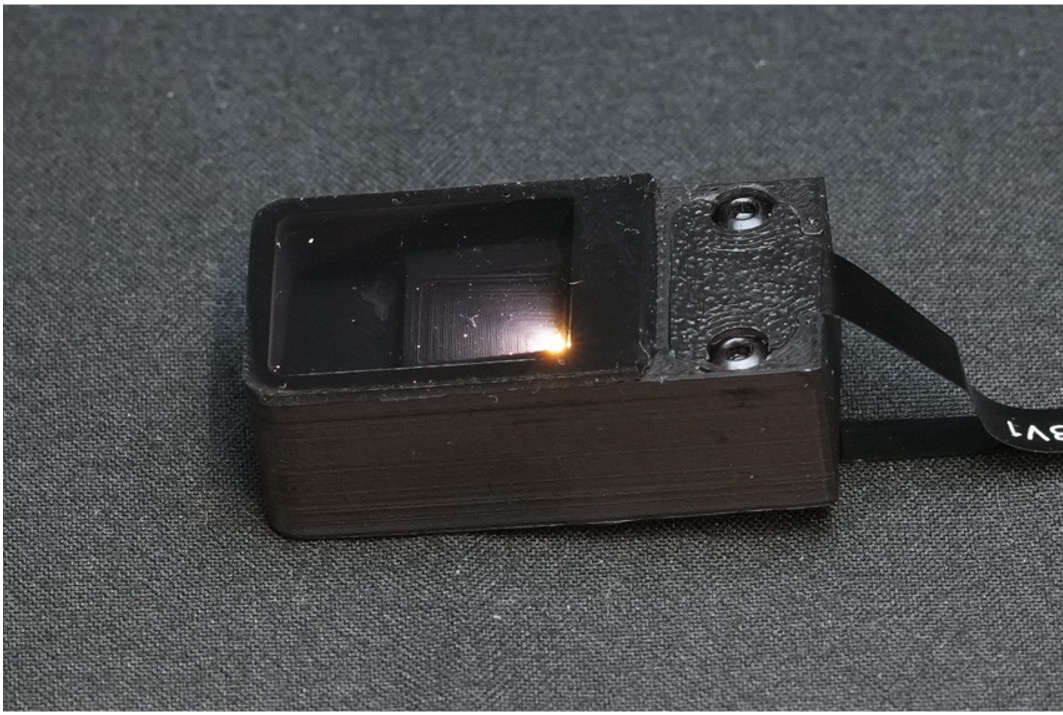}}
& \multirow{\rownumenv}{*}{\includegraphics[width = \widthenv]{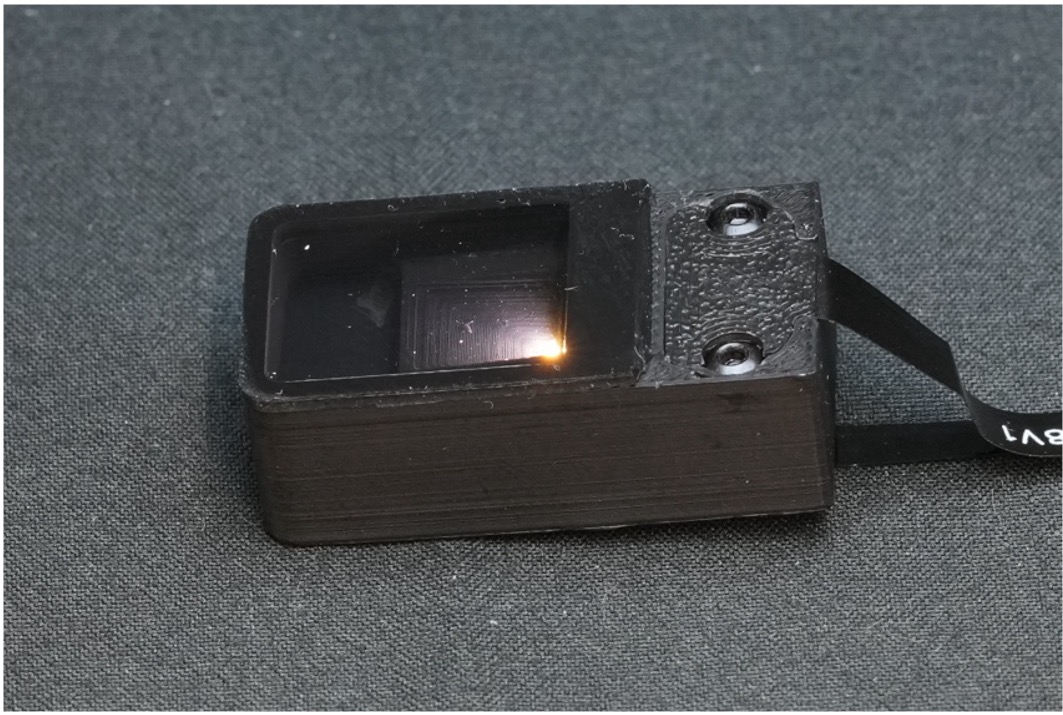}}
& \multirow{\rownumenv}{*}{\includegraphics[width = \widthenv]{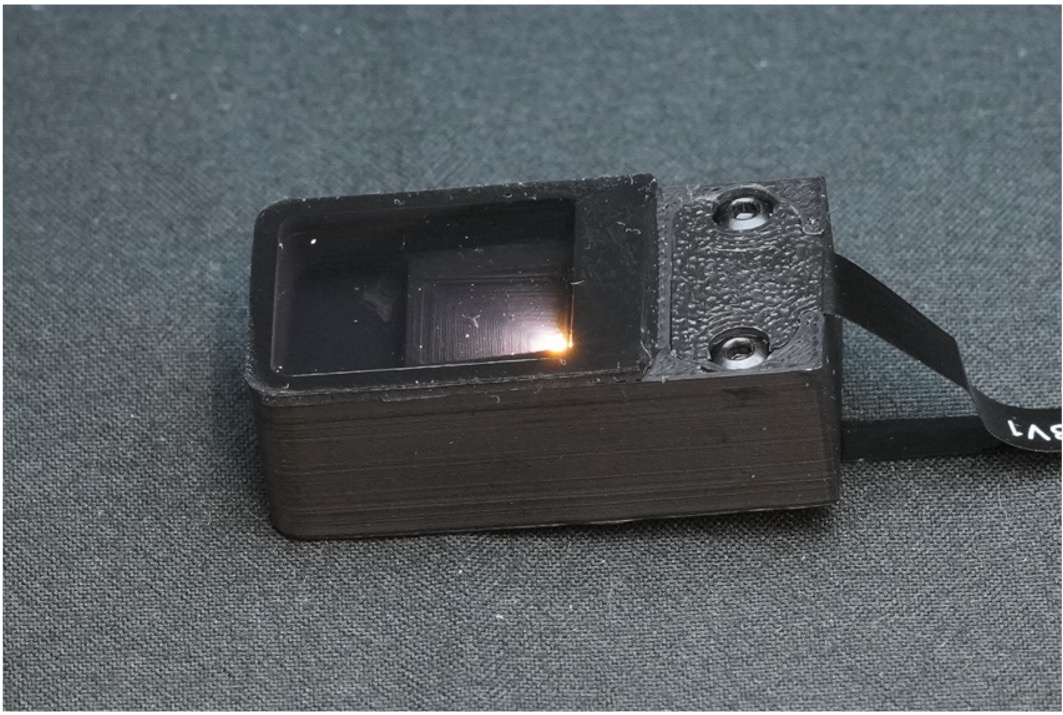}}
& \multirow{\rownumenv}{*}{\includegraphics[width = \widthenv]{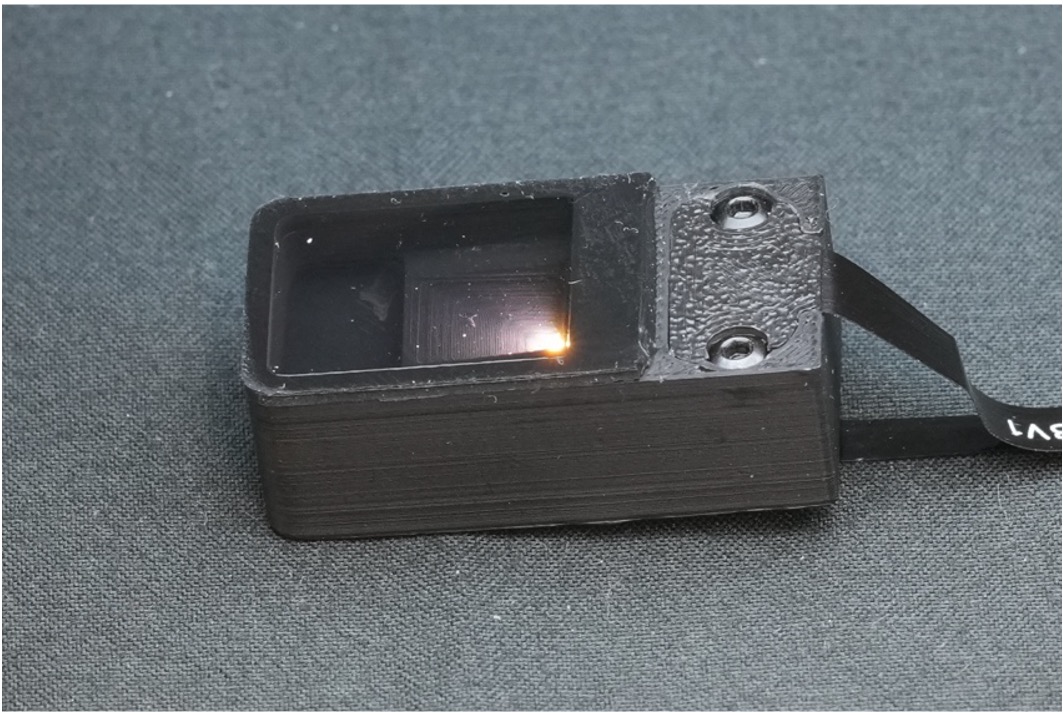}}
\\ \\ \\
\midrule
Non-contact image mean & 0.41 & 1.00 & 1.79 & 4.26 & 1.28 & 1.52 & 2.40 & 2.94 & 4.69 & 6.90
\\
Non-contact image std & 0.49 & 0.33 & 0.97 & 3.45 & 0.47 & 0.64 & 1.09 & 1.61 & 2.83 & 4.14
\\
\multicolumn{1}{l}{\multirow{\rownumber}{*}{Non-contact raw image}}
& \multirow{\rownumber}{*}{\includegraphics[angle=90, width = \lengthref]{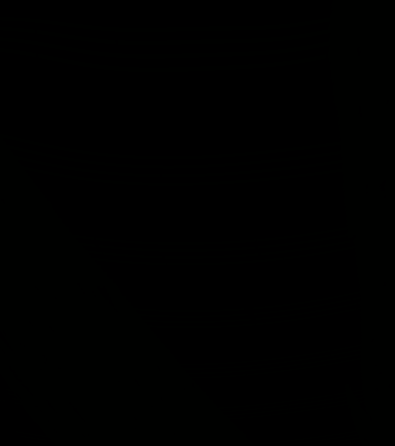}}
& \multirow{\rownumber}{*}{\includegraphics[angle=90, width = \lengthref]{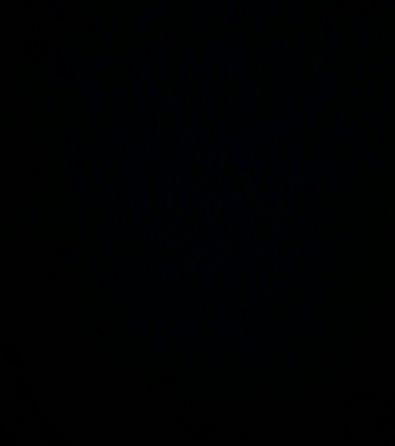}}
& \multirow{\rownumber}{*}{\includegraphics[angle=90, width = \lengthref]{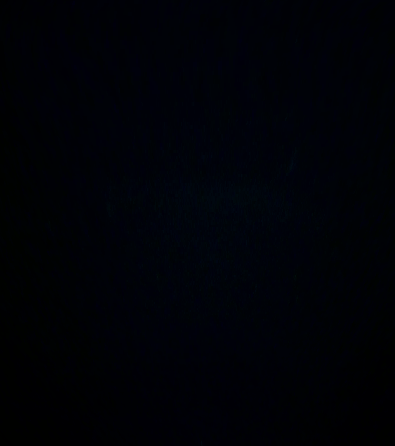}}
& \multirow{\rownumber}{*}{\includegraphics[angle=90, width = \lengthref]{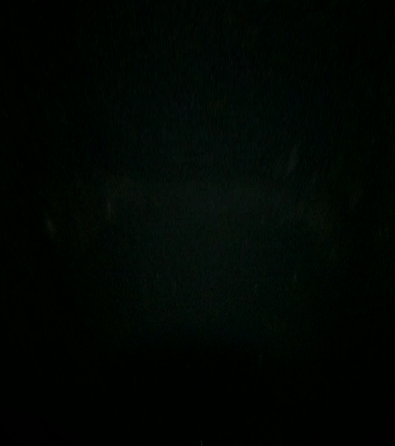}}
& \multirow{\rownumber}{*}{\includegraphics[angle=90, width = \lengthref]{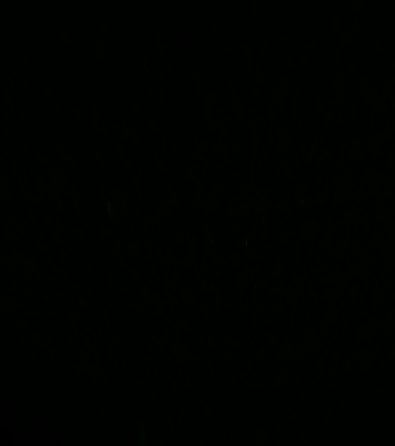}}
& \multirow{\rownumber}{*}{\includegraphics[angle=90, width = \lengthref]{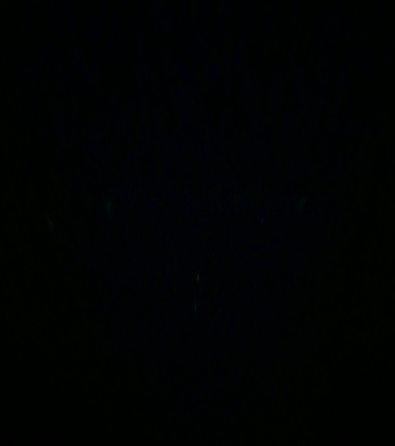}}
& \multirow{\rownumber}{*}{\includegraphics[angle=90, width = \lengthref]{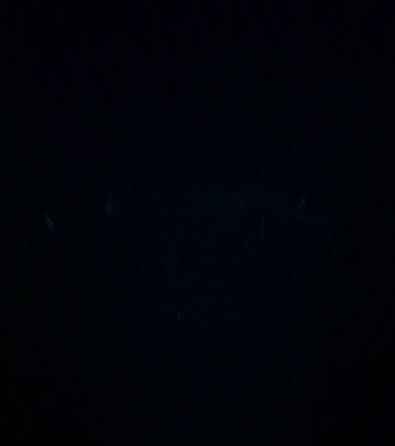}}
& \multirow{\rownumber}{*}{\includegraphics[angle=90, width = \lengthref]{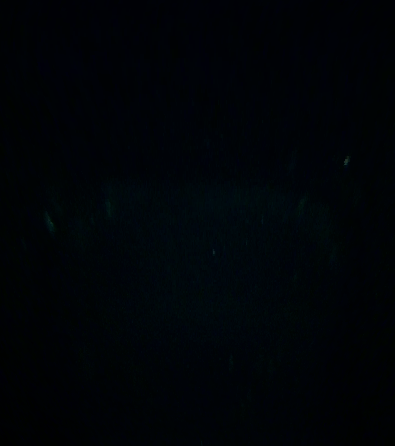}}
& \multirow{\rownumber}{*}{\includegraphics[angle=90, width = \lengthref]{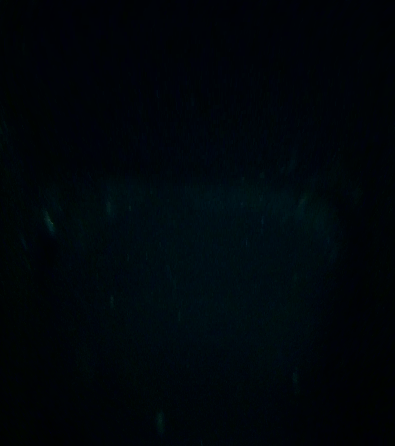}}
& \multirow{\rownumber}{*}{\includegraphics[angle=90, width = \lengthref]{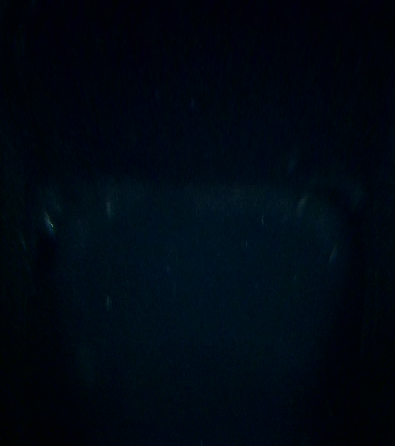}}
\\ \\ \\ \\
\midrule
\multicolumn{1}{l}{\multirow{\rownumber}{*}{Contact raw image}}
& \multirow{\rownumber}{*}{\includegraphics[angle=90, width = \lengthref]{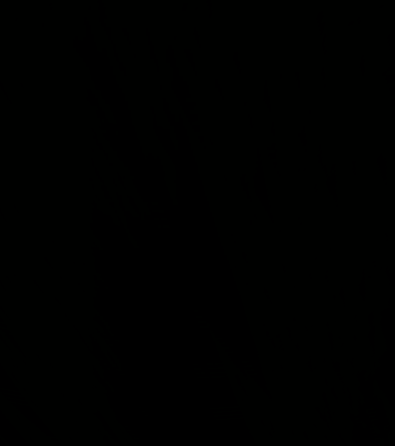}}
& \multirow{\rownumber}{*}{\includegraphics[angle=90, width = \lengthref]{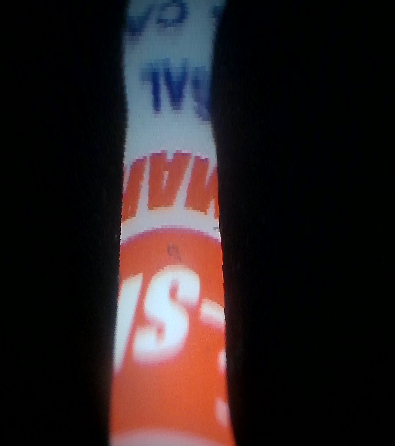}}
& \multirow{\rownumber}{*}{\includegraphics[angle=90, width = \lengthref]{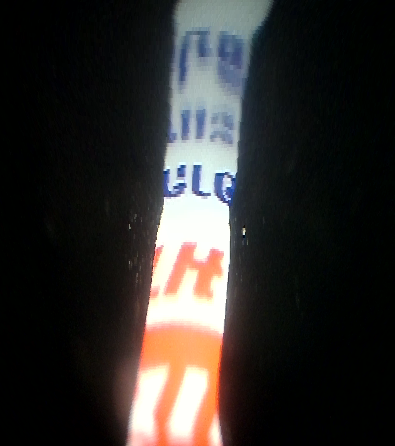}}
& \multirow{\rownumber}{*}{\includegraphics[angle=90, width = \lengthref]{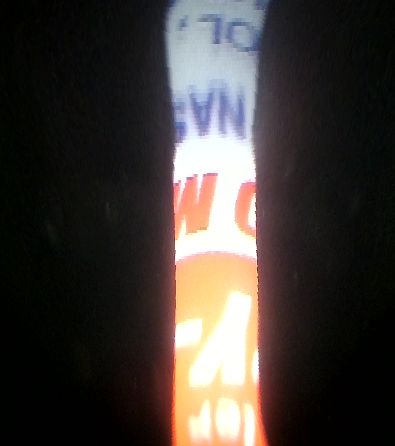}}
& \multirow{\rownumber}{*}{\includegraphics[angle=90, width = \lengthref]{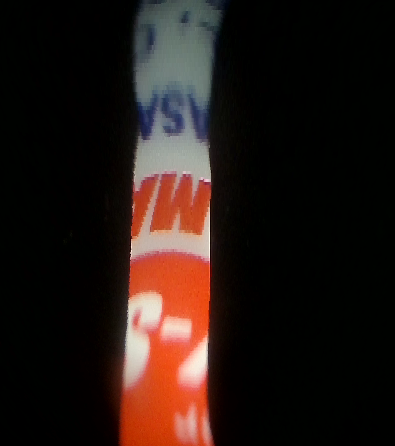}}
& \multirow{\rownumber}{*}{\includegraphics[angle=90, width = \lengthref]{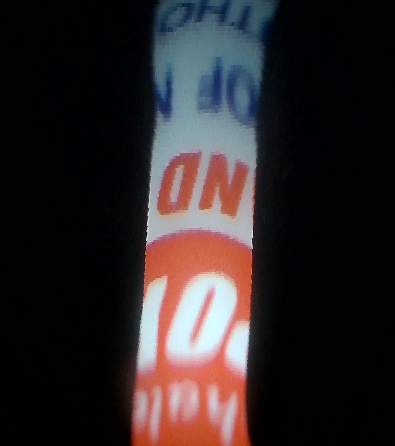}}
& \multirow{\rownumber}{*}{\includegraphics[angle=90, width = \lengthref]{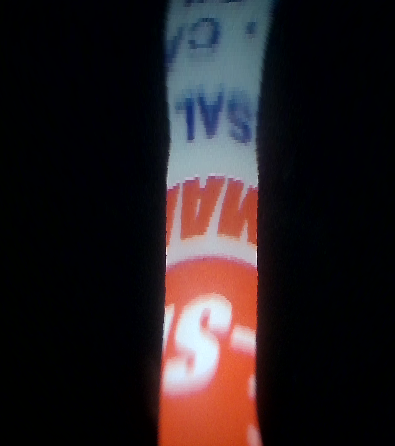}}
& \multirow{\rownumber}{*}{\includegraphics[angle=90, width = \lengthref]{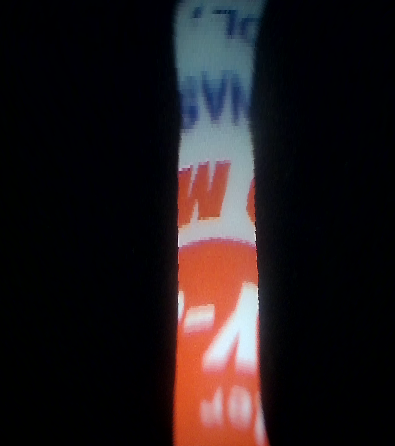}}
& \multirow{\rownumber}{*}{\includegraphics[angle=90, width = \lengthref]{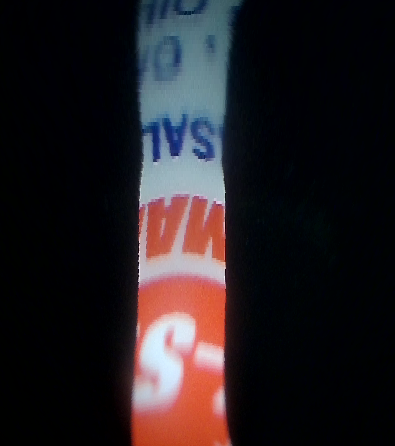}}
& \multirow{\rownumber}{*}{\includegraphics[angle=90, width = \lengthref]{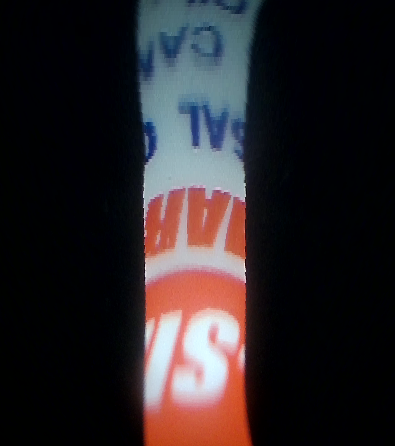}}
\\ \\ \\ \\
\multicolumn{1}{l}{\multirow{\rownumber}{*}{Contact segmentation}}
& \multirow{\rownumber}{*}{\includegraphics[angle=90, width = \lengthref]{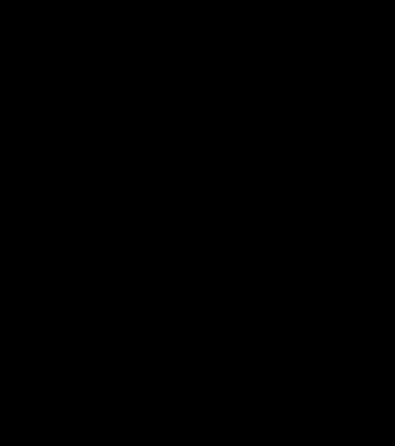}}
& \multirow{\rownumber}{*}{\includegraphics[angle=90, width = \lengthref]{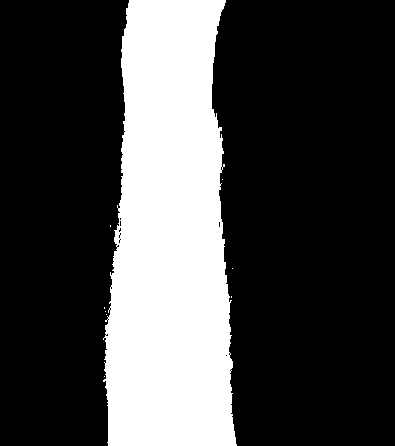}}
& \multirow{\rownumber}{*}{\includegraphics[angle=90, width = \lengthref]{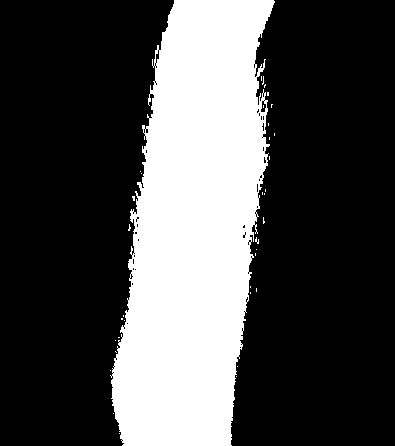}}
& \multirow{\rownumber}{*}{\includegraphics[angle=90, width = \lengthref]{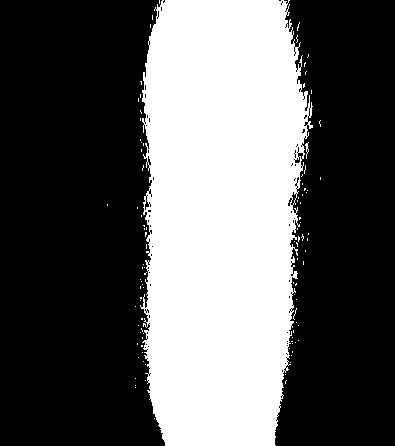}}
& \multirow{\rownumber}{*}{\includegraphics[angle=90, width = \lengthref]{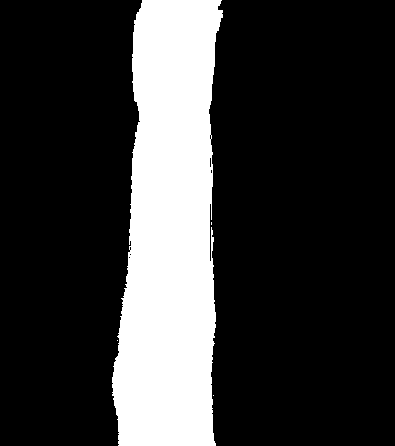}}
& \multirow{\rownumber}{*}{\includegraphics[angle=90, width = \lengthref]{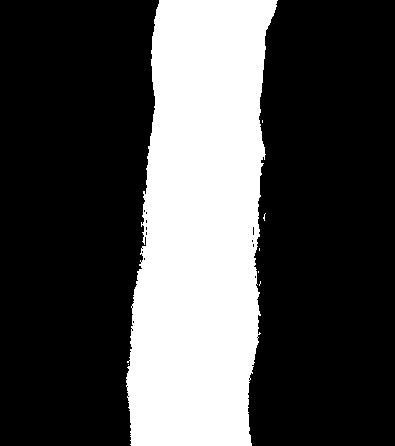}}
& \multirow{\rownumber}{*}{\includegraphics[angle=90, width = \lengthref]{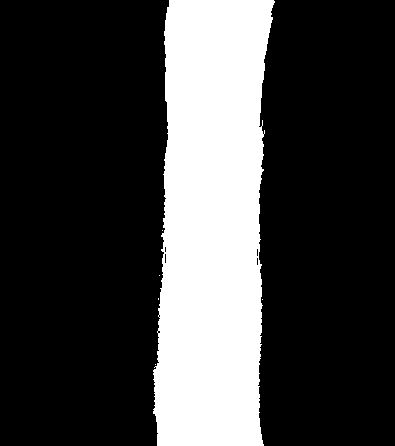}}
& \multirow{\rownumber}{*}{\includegraphics[angle=90, width = \lengthref]{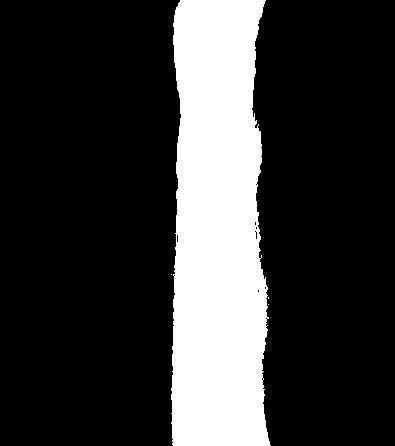}}
& \multirow{\rownumber}{*}{\includegraphics[angle=90, width = \lengthref]{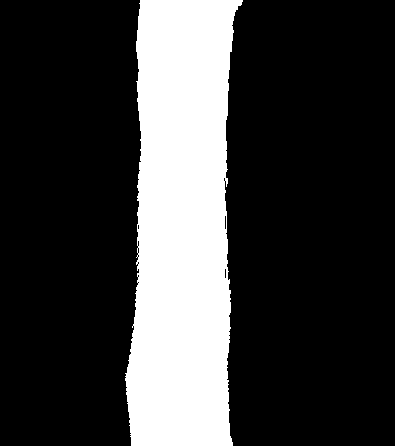}}
& \multirow{\rownumber}{*}{\includegraphics[angle=90, width = \lengthref]{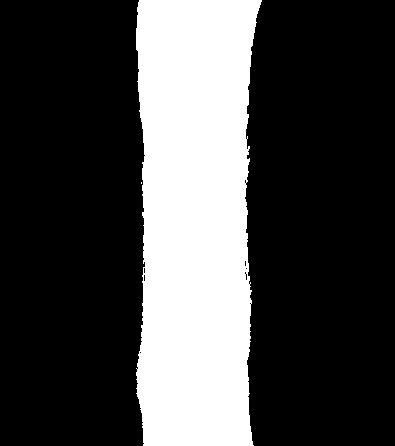}}
\\ \\ \\ \\
\bottomrule
\end{tabular}
\label{tab:light_blocking}
\end{table*}

\subsection{Contact Segmentation}
\label{subsec:contact_segmentation}

\begin{figure}[t]
\centering
\includegraphics[width=\linewidth]{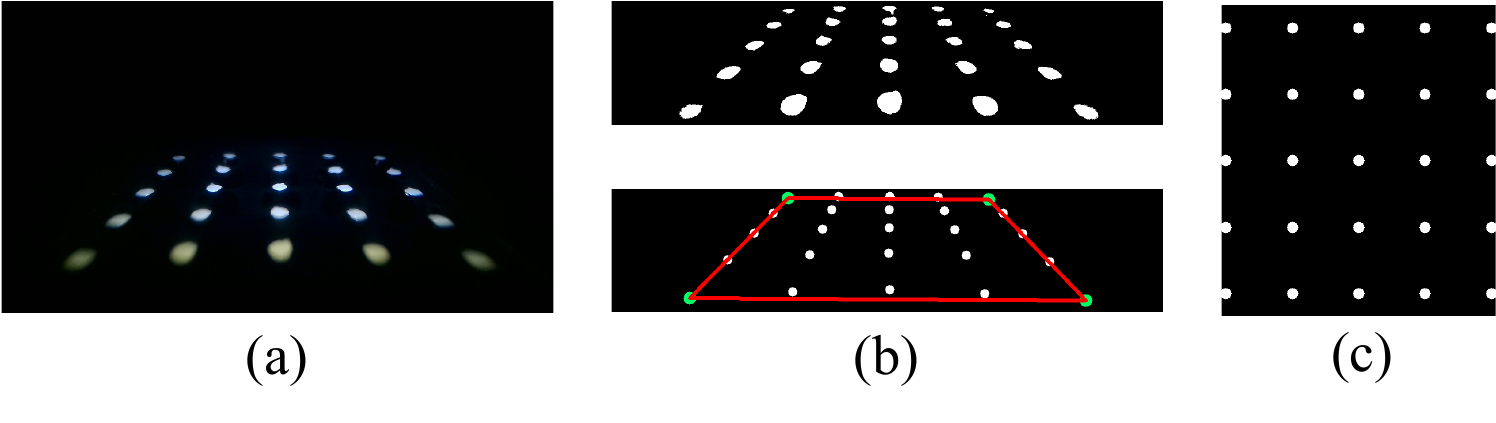}
\vspace{-0.8cm}
\caption{Camera calibration for \sensor.
(a) \sensor captures imprints produced by a calibration tool with a $5\times5$ array of cylinders.
(b) The imprints are segmented by thresholding and their center pixels are detected.
(c) The final rectified and cropped image.
}
\vspace{-0.4cm}
\label{fig:calibration}
\end{figure}

\textbf{Camera operation and sensor calibration.}
We disable auto-exposure and operate the camera at a fixed exposure time of $20~\text{ms}$.
This is crucial because auto-exposure would brighten \sensor's predominantly dark background, amplifying sensor noise and potentially saturating bright contact regions.
For robotic use, we further calibrate a mapping from image pixels to physical locations on the touching surface.
Following 9DTact~\cite{lin20239dtact}, we use a calibration tool with a $5\times5$ array of cylindrical bumps at $3~\text{mm}$ spacing.
We press the tool onto the surface and capture the resulting imprint pattern (Fig.~\ref{fig:calibration}(a)).
We segment the imprints via thresholding and extract their center pixels (Fig.~\ref{fig:calibration}(b)).
We then use the four outermost corner imprints to estimate the average pixel spacing and define the target rectified grid.
Finally, we compute a warping map from the detected centers to the rectified grid, and apply it to rectify and crop each raw image into a top-down view (Fig.~\ref{fig:calibration}(c)).

\textbf{Segmentation algorithm.}
The optical design of \sensor naturally produces a high-contrast representation of contact: non-contact regions remain near-black, while true contact results in localized brightness increases.
Building on this property, our segmentation algorithm adopts a simple yet highly robust frame-differencing strategy.
Before operation, \sensor records $N$ no-contact frames and computes their pixel-wise average as a reference image $I_{\mathrm{ref}}$.
Given an incoming frame $I$, we compute a difference image $I_{\mathrm{diff}} = I_{\mathrm{raw}} - I_{\mathrm{ref}}
$, in which contact pixels appear as positive brightness changes.

To separate true contact from sensor noise, we apply a multi-condition brightness consistency test.
A pixel at location $(x,y)$ is classified as contact if any of the following holds:
(1) the mean RGB increase exceeds $t_0$;
(2) at least one channel exceeds $t_1$;
(3) at least two channels exceed $t_2$; or
(4) all three channels exceed $t_3$.
These complementary criteria capture both strong changes and weaker but consistent increases produced by contact.
In our implementation, we use $N=10$ and ${t_0,t_1,t_2,t_3}={25,20,30,40}$.

\section{Experiments and Results}
In this section, we conduct experiments to mainly answer the following questions:
1) Can \sensor achieve the intended light-suppression behavior and produce natural high-contrast contact images?
2) Can \sensor perform robust deformation-independent contact segmentation across diverse materials, appearances, contact forces, and ambient illumination?
3) What new manipulation capabilities does \sensor unlock when equipped on robots?

% We conduct experiments to answer the following questions:
% 1) Does \sensor achieve the intended light-suppression behavior, keeping non-contact regions dark while preserving natural contact appearance?
% 2) Can \sensor quantitatively segment contact regions under known contact geometry?
% 3) Does \sensor generalize to deformation-independent contact sensing across diverse materials, appearances, contact forces, and lighting conditions?
% 4) What robotic capabilities are enabled by \sensor when conventional deformation-based tactile sensors fail?

\begin{figure*}[t]
\centering
\includegraphics[width=\linewidth]{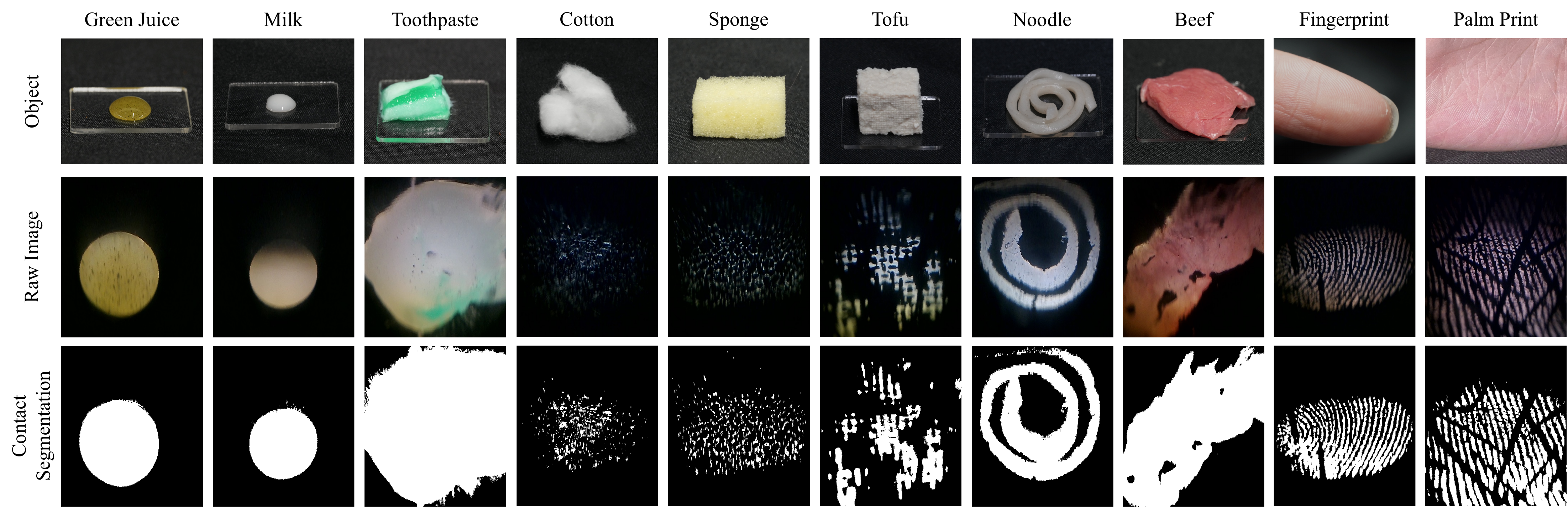}
\vspace{-0.6cm}
\caption{
\sensor achieves robust contact segmentation with objects that make extremely light contact without producing macroscopic surface deformation, such as liquids, semi-liquids, and ultra-soft materials.
}
\vspace{-0.4cm}
\label{fig:segmentation_liquid_soft}
\end{figure*}

\subsection{Light Suppression at Non-Contact Regions}
We first evaluate whether \sensor achieves the intended light-suppression behavior under varying levels of internal illumination and external ambient brightness.
We measure the brightness at the touching surface using a digital light meter. 
For each condition, we record (1) raw images without contact (non-contact images) and (2) raw images with a cylindrical object touching the sensing surface (contact images).
We report the mean and standard deviation of the non-contact images in Table~\ref{tab:light_blocking}, along with representative raw and segmented results.

\textbf{Internal illumination.}
We first test \sensor in a dark environment and gradually increase the LED voltage.
When the LED is off, both contact and non-contact images remain near-black as expected.
As LED brightness increases to $430~\text{Lux}$, the contact images begin to reveal clear visual appearance of the object while the non-contact regions remain extremely dark (mean gray value $=1.0$).
Further increasing the LED brightness (e.g., $1050$ or $1670~\text{Lux}$) causes only slight increases in non-contact intensity.
However, these higher illumination levels overexpose the contact regions and degrade object appearance.
To maintain near-black non-contact regions and capture clear object appearance, we set $430~\text{Lux}$ (corresponding to a $3.4~\text{V}$ driving voltage) as the default internal illumination for \sensor.

\textbf{External illumination.}
Using the default LED setting, we then increase external illumination by gradually brightening an external lamp.
As shown in Table~\ref{tab:light_blocking}, \sensor maintains its non-contact pixels near-black (mean gray value $<3$) even under strong ambient lighting up to $2010~\text{Lux}$, which is far above typical indoor brightness ($<1000~\text{Lux}$).
At extremely bright illumination conditions ($3520~\text{Lux}$), the non-contact image becomes slightly brighter (mean gray value $=6.90$) due to imperfect absorption on the internal surfaces of the sensor, as discussed in Section~\ref{subsec:sensor_design}.
Nonetheless, contact regions remain clearly visible and segmentation remains robust.

These results demonstrate that \sensor consistently suppresses both external and internal light at non-contact regions while preserving natural appearance in contact regions.
This behavior directly validates our optical design and highlights a key distinction from TIR-based sensors~\cite{han2005low, shimonomura2016robotic, zhang2023tirgel, sun2025soft}, which fail to distinguish contact regions under bright environments.

\subsection{Deformation-Independent Contact Segmentation}

\textbf{Extremely delicate contact sensing.}
Figure~\ref{fig:segmentation_liquid_soft} evaluates \sensor's ability to segment contact for liquids (green juice, milk), semi-liquids (toothpaste), and ultra-soft materials (cotton, sponge, tofu, noodle, beef, fingertip, palm) under bright indoor illumination.
Across all trials, objects are placed or gently touched onto the sensor without intentional pressing, resulting in minimal or no macroscopic surface deformation.
Despite the absence of indentation cues, \sensor directly identifies the contact region from light intensity changes in the captured image, as its optical light-suppression mechanism produces clean, pixel-level segmentation even for visually complex materials.
By contrast, deformation-based tactile sensors, including 9DTact~\cite{lin20239dtact}, GelSight~\cite{yuan2017gelsight}, and DelTact~\cite{zhang2022deltact}, do not yield reliable contact detection in this near-zero-force regime, since their measurements fundamentally rely on measurable deformation.
Moreover, \sensor can detect contact under effectively zero applied force, as shown by the thin film hanging upside down on the downward-facing \sensor in Fig.~\ref{fig:teaser}; this further indicates that \sensor does not require a minimum force threshold to trigger contact perception.

\textbf{Sensing across materials.}
\begin{figure}[t]
\centering
\includegraphics[width=\linewidth]{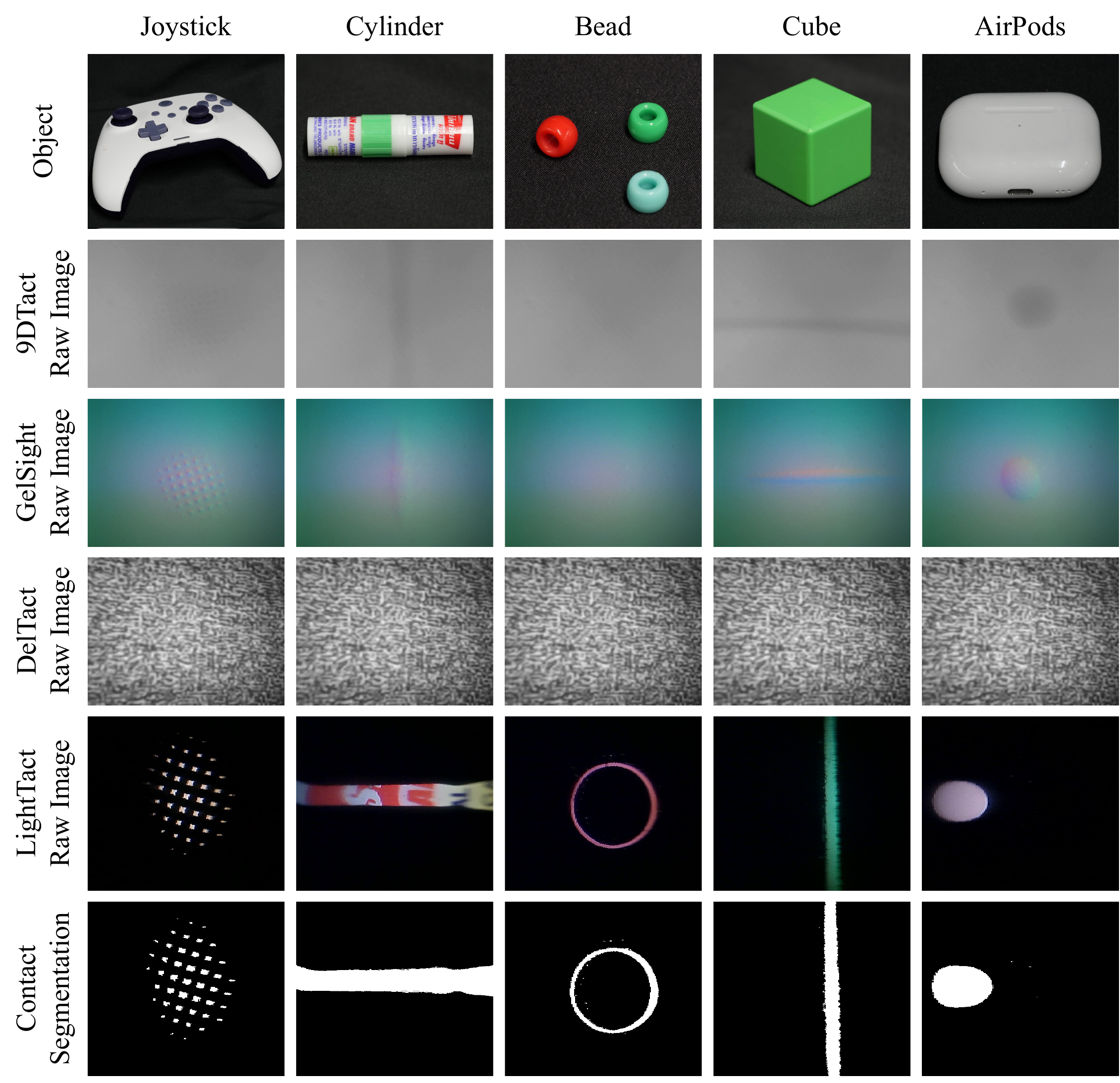}
\vspace{-0.6cm}
\caption{
\sensor reliably senses both light and firm contact from rigid objects while preserving their visual appearance.
In contrast, deformation-based tactile sensors fail under light contact, as the bead example illustrated.}
\vspace{-0.4cm}
\label{fig:segmentation_hard}
\end{figure}
Beyond liquids and soft materials, \sensor also provides robust visual--tactile sensing for rigid objects, as shown in Fig.~\ref{fig:segmentation_hard} (textured joystick surface, complex-pattern cylinder, beads, cube, AirPods).
Across both gentle placement and firmer pressing, \sensor consistently produces accurate contact segmentation while preserving natural visual appearance in the raw image.
In contrast, deformation-based VBTSs become unreliable with rigid bodies when contact does not induce sufficient deformation (e.g., the bead case).
Overall, these results show that \sensor performs accurate, pixel-level contact segmentation across liquids, semi-liquids, ultra-soft materials, and rigid objects.
Such robust coverage across materials, appearances, and lighting is difficult to achieve with both TIR-based and deformation-based VBTSs.

\subsection{Robotic Capabilities Unlocked by \sensor}
To showcase the manipulation abilities enabled by \sensor's direct, deformation-independent contact perception, we conduct a series of experiments on a UFACTORY xArm~7 robotic arm equipped with one or two \sensor modules mounted on its gripper.
These demonstrations highlight two classes of capabilities:
\begin{itemize}
\item \textbf{Light contact sensing} for interaction with liquids, semi-liquids, and ultra-soft materials;
\item \textbf{Multimodal reasoning} for fine-grained manipulation by directly prompting \sensor images to vision-language models (VLMs).
\end{itemize}
For all experiments, we additionally compare \sensor with a force-torque (FT) sensor and representative vision-based tactile sensors that are either open-source or commercially available, including 9DTact~\cite{lin20239dtact}, GelSight-Mini~\cite{gelsight-mini2022}, and DelTact~\cite{zhang2022deltact}. We refer to these as \emph{baseline sensors} below.

\subsubsection{Light contact sensing of liquids, semi-liquids, and soft materials}
When a robot approaches an object for grasping or manipulation, the wrist-camera view is often occluded by the end-effector and the object, making tactile sensing essential for safe, closed-loop interaction.
However, most VBTSs rely on contact-induced deformation and thus provide weak or unreliable signals when contact produces little-to-no indentation.
In these regimes, \sensor remains effective: it detects contact at the gripper via a deformation-independent optical principle.

\begin{figure}[t]
\centering
\includegraphics[width=\linewidth]{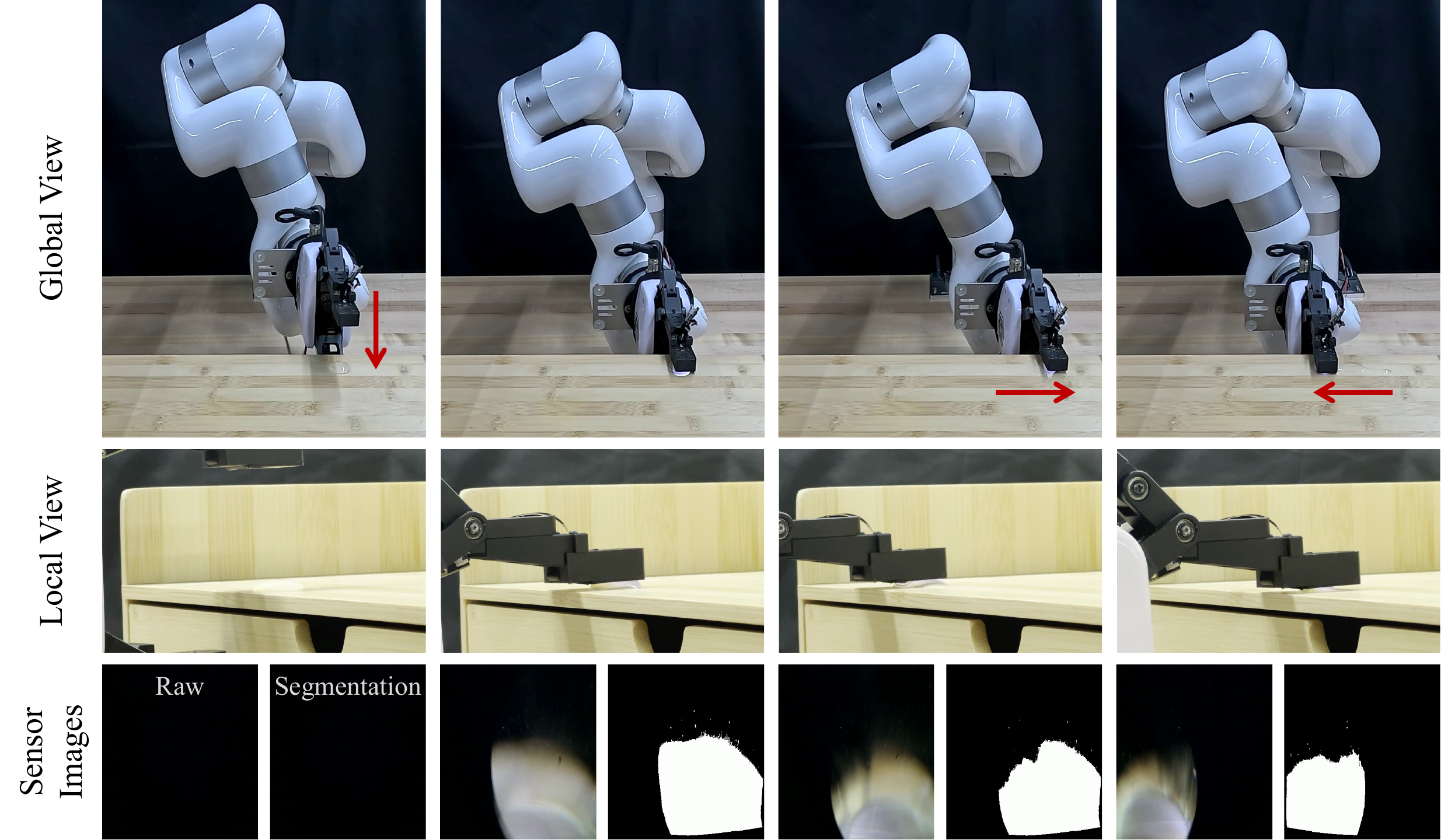}
\vspace{-0.6cm}
\caption{
\sensor detects contact with water and segments the contact region, enabling the robot to spread water while maintaining contact and avoiding collision with the surface.
}
% \vspace{-0.2cm}
\label{fig:water_spreading}
\end{figure}

\begin{figure}[t]
\centering
\includegraphics[width=\linewidth]{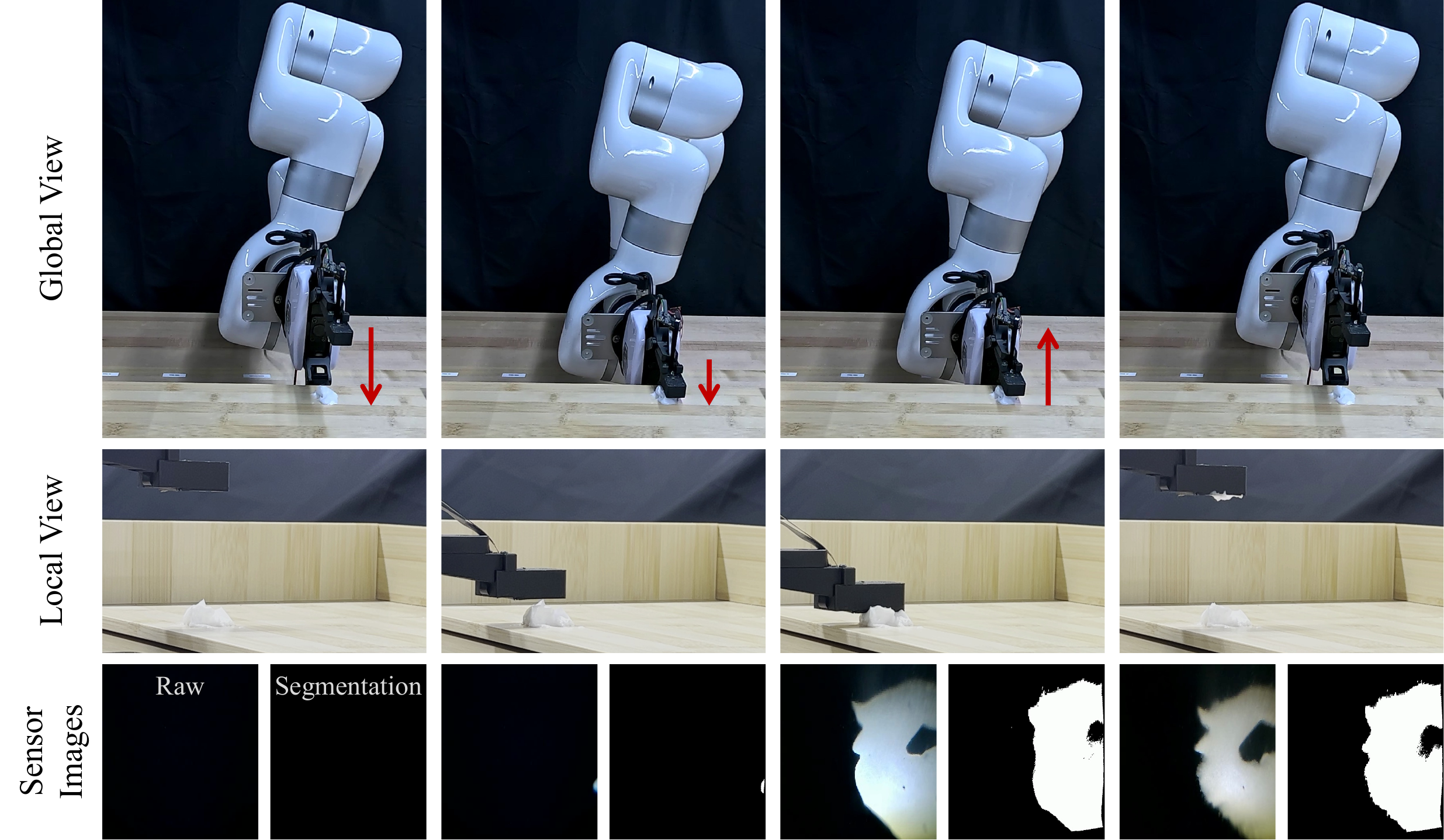}
\vspace{-0.6cm}
\caption{
Using \sensor, the robot can dip facial cream using a contact-coverage criterion.
}
\vspace{-0.4cm}
\label{fig:cream_dipping}
\end{figure}

\textbf{Liquid spreading.}
As shown in Fig.~\ref{fig:water_spreading}, a small amount of water is placed on a flat cabinet, and the robot orients \sensor downward.
The gripper descends until contact is detected, then performs lateral sweeps while continuously adjusting its height to maintain contact with the water and avoid colliding with the surface.
We regulate the end-effector height using a PD controller to maintain approximately $50\%$ contact coverage.
This closed-loop policy allows the robot to spread the water uniformly.

\textbf{Semi-liquid dipping.}
As shown in Fig.~\ref{fig:cream_dipping}, to collect facial cream, the robot moves its fingertip toward the cream, reduces its approach speed once contact is detected, and stops when more than $50\%$ of the sensing surface is in contact.
Notably, \sensor continues to detect the cream reliably during lifting, where essentially no pressure is applied on the sensing surface.

\begin{figure}[t]
\centering
\includegraphics[width=\linewidth]{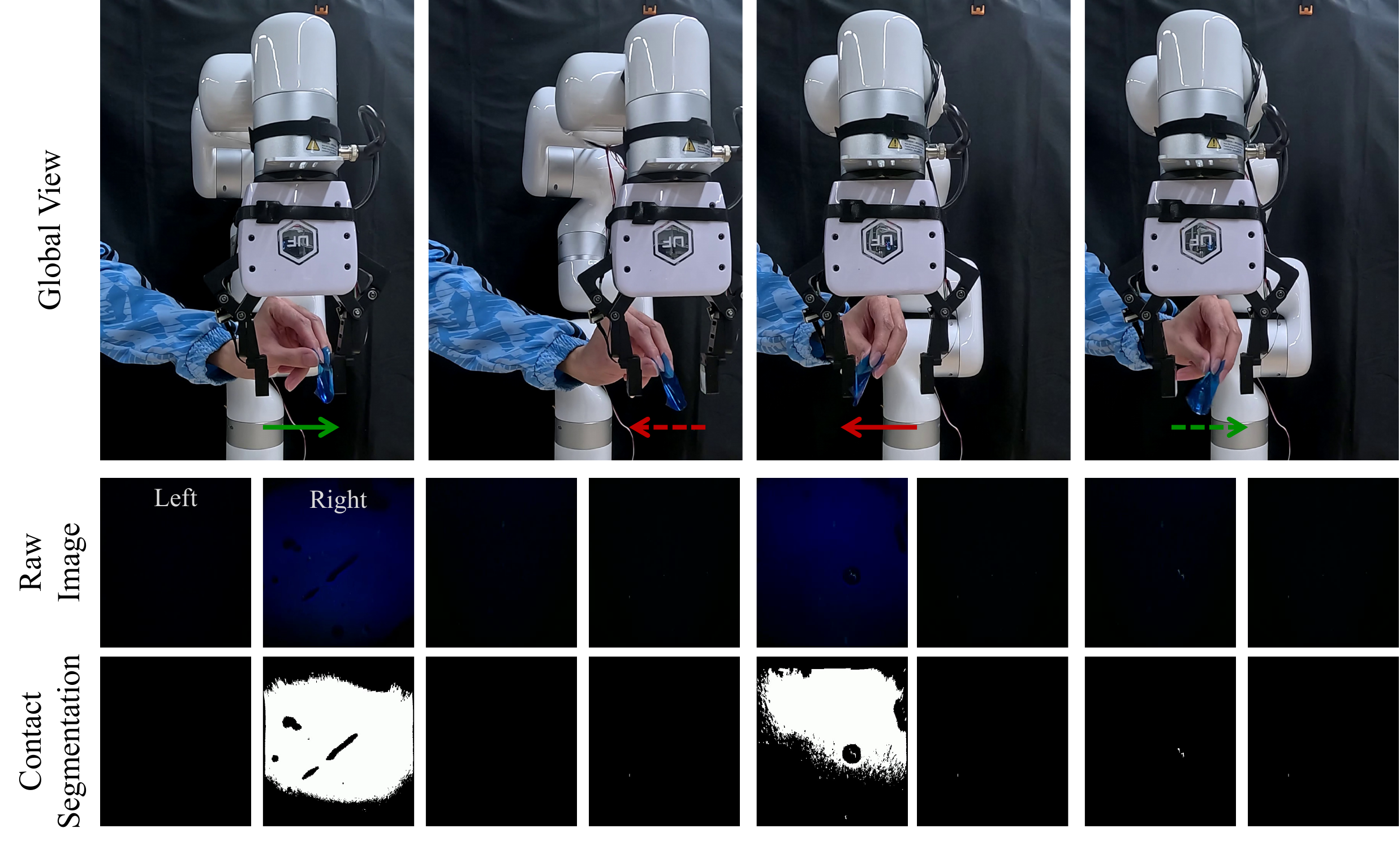}
\vspace{-0.6cm}
\caption{
Two \sensor sensors enable responsive interaction with an ultra-thin film.
Solid arrows indicate the end-effector motion commanded by contact on the corresponding sensor; dotted arrows indicate return to the centered position when no contact is detected.
\vspace{-0.4cm}}
\label{fig:thin_film}
\end{figure}
\textbf{Extremely soft materials.}
As shown in Fig.~\ref{fig:thin_film}, a human holds a small piece of food film that is ultra-soft, thin ($0.1~\text{mm}$), and light ($0.05~\text{g}$), and lightly touches it to a gripper equipped with two \sensor sensors.
When the right sensor detects contact, the robot moves right; when the left sensor detects contact, it moves left; and when no contact is detected, it returns to a centered position.
This behavior illustrates that \sensor can reliably sense minimal, delicate contacts and support responsive human-robot interaction.

Across all three tasks, the baseline sensors fail to produce usable signals because the contacts are too light to generate measurable forces or macroscopic deformation.
These results underscore \sensor's unique ability to enable robotic behaviors that require precise sensing of extremely light contact.

\subsubsection{Multimodal sensing and VLM-guided fine-grained manipulation}
Vision-language models (VLMs) are emerging as powerful general-purpose reasoning engines for robotics~\cite{gao2024physically, team2025gemini}.
However, conventional tactile inputs, either sparse force signals or deformation-based tactile images, provide limited appearance information and are often out-of-distribution for VLMs, making them difficult to interpret directly for manipulation~\cite{yu2025demonstrating, huang2025tactile, cheng2025omnivtla}.
In contrast, \sensor produces high-resolution, appearance-preserving images in which contact geometry and local object appearance are naturally aligned, making them directly compatible with VLM-based reasoning.

\textbf{Resistor grasping and resistance reasoning.}
To demonstrate this capability, the robot is tasked with grasping and sorting resistors by their resistance values.
The gripper closes until one \sensor detects more than 100 contact pixels, yielding a stable and gentle grasp.
We crop the raw image around the contact region and prompt a VLM (GPT-5 Pro in our implementation) to infer the resistance value from the color bands.
The robot then places the resistor into the corresponding cup.
We evaluate this pipeline over 20 trials with 5 resistors and achieve 16 successful sorts ($80\%$ success rate).
Most failures arise from the VLM confusing visually similar band colors (e.g., red vs.\ brown vs.\ orange).

\begin{figure}[ht]
\centering
\includegraphics[width=\linewidth]{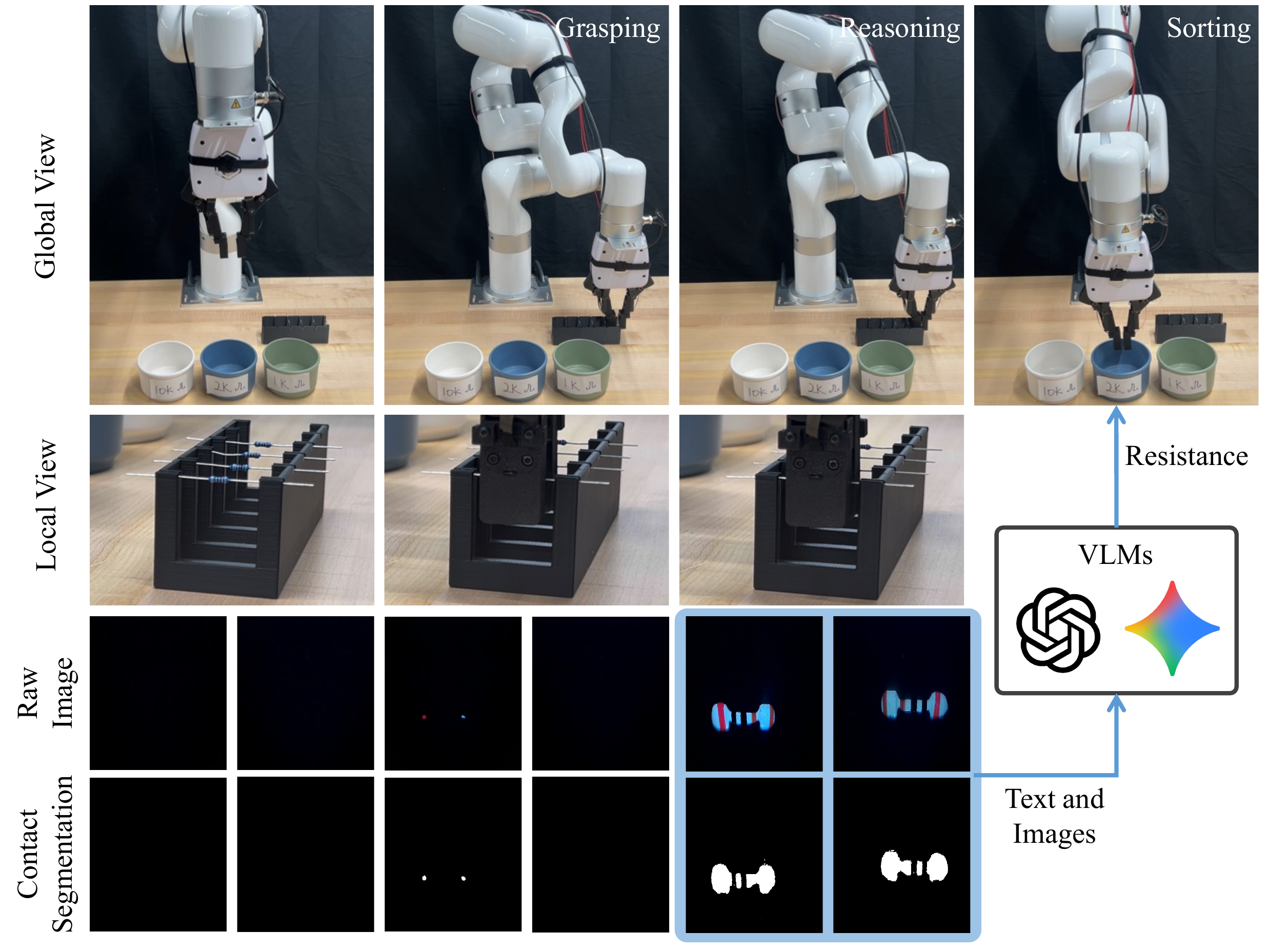}
\vspace{-0.6cm}
\caption{Beyond pixel-level contact sensing, \sensor preserves the natural appearance of the contacting surface, enabling VLMs to reason directly over its images.
In this task, a VLM infers resistor values from color bands to guide robotic sorting.
\vspace{-0.4cm}}
\label{fig:resistor_sorting}
\end{figure}

In contrast, the baseline sensors do not provide interpretable appearance cues; therefore cannot support this VLM-guided sorting pipeline.
We also compare against wrist-camera images: the VLM fails consistently because the resistor occupies a small fraction of the image and the background is cluttered.
Overall, these results demonstrate that \sensor can provide VLM-readable inputs without fine-tuning, enabling robots to obtain high-level task decisions directly through prompting.

\section{Ablation Study for Sensor Design}
We evaluate ablation studies for several key design and fabrication choices introduced in
Section~\ref{sec:methodology}, with full results reported in Appendix~\ref{app:design_ablation}.
\begin{itemize}
    \item Black paint on the shell inner walls
    \item Acrylic mold for casting the black gel
    \item Edge indentation on the shell outer walls
    % \item Transparent gel geometry (1\,mm, 2\,mm, $45^\circ$)
\end{itemize}

\section{Conclusion}
We presented \sensor, a compact visual-tactile fingertip sensor that enables deformation-independent, optics-based contact sensing.
The core idea is an ambient-blocking optical configuration that suppresses both external light and internal illumination at non-contact regions, while transmitting only the scattered light generated at true contacts.
This yields high-contrast raw images with near-black non-contact pixels and appearance-preserving contact pixels, making pixel-level contact segmentation simple and robust.
Across systematic evaluations, \sensor maintains reliable contact segmentation over diverse materials (liquids, semi-liquids, ultra-soft materials, and rigid objects) and under strong ambient lighting.
When integrated on a robotic arm, \sensor enables manipulation behaviors driven by extremely light contact, including water spreading, facial-cream dipping, and thin-film interaction, where force-based sensing and deformation-based VBTSs fail. This is promising for applications such as delicate handling of cosmetics, food, and biomedical samples, operating on wet surfaces, and enabling safe human-robot interaction.
Moreover, \sensor provides spatially aligned visual-tactile images that can be directly prompted to vision-language models for fine-grained reasoning, demonstrated by resistor value inference for robotic sorting.

\clearpage
\section*{Acknowledgments}
We would like to thank Yuxiang Yang for his insightful discussions throughout the project.

%% Use plainnat to work nicely with natbib. 
\newcommand{\betweenfs}{\fontsize{8pt}{10.2pt}\selectfont}
{
% \footnotesize
% \betweenfs
\bibliographystyle{IEEEtran}
\bibliography{main}
}

\clearpage
\section*{Appendix}

\subsection{Camera Placement under $\theta_{tv}<2\theta_c$}
\label{app:camera_placement}
In this section, we discuss our motivation to design the angle between the touching and viewing surfaces $\theta_{tv}$.
The optical objective is to prevent the camera from seeing external light from non-contact regions of the touching surface.
To achieve this, we choose $\theta_{tv}=\frac{\pi}{2}>2\theta_c$ for \sensor (Section.~\ref{sec:methodology}) so that these rays meet TIR when they reach the viewing surface and cannot transmit through it.

\begin{figure}[ht]
\centering
\includegraphics[width=\linewidth]{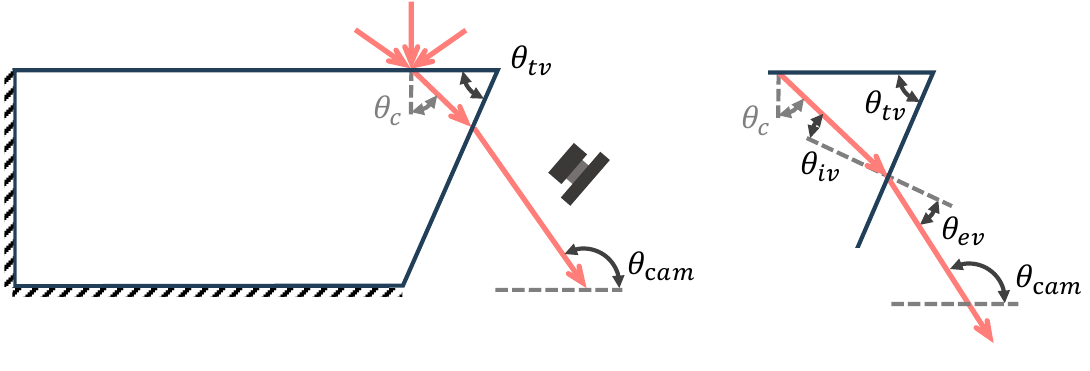}
\vspace{-0.6cm}
\caption{
Optical analysis for the transparent medium under $\theta_{tv}<2\theta_c$.
External light can transmit through the viewing surface so the camera should be placed outside the areas where external light can arrive.
}
\label{fig_app:tv_angle}
\end{figure}

Another more relaxed design is to allow these rays to transmit through the viewing surface but with the camera outside the ray field.
This choice allows $\theta_{tv}<2\theta_c$ as illustrated in Fig.~\ref{fig_app:tv_angle}.
In this situation, ambient light first transmits through the touching surface and reach the viewing surface.
The incident angle has a maximal value of $\theta_{iv}=\theta_{tv}-\theta_c$.
And the corresponding minimal exit angle is
\begin{equation}
\begin{split}
\theta_{ev} &=f(\theta_{iv}) \\
&= \sin^{-1}(\frac{n_m}{n_a} \sin\theta_{iv})
\label{exit_angle}
\end{split}
\end{equation}
where $n_m$ and $n_a$ are the refractive index of the transparent medium (around $1.45$) and the air (around $1.0$).
Therefore, the angle between the most slant ray and the horizontal line is
\begin{equation}
\begin{split}
\theta_{cam} &= \frac{\pi}{2}+\theta_{tv}-\theta_{ev} \\
&= \frac{\pi}{2}+\theta_{tv}-\sin^{-1}(\frac{n_m}{n_a} \sin\theta_{iv}) \\
&= \frac{\pi}{2}+\theta_{tv}-\sin^{-1}(\frac{n_m}{n_a} \sin(\theta_{tv}-\theta_c)) \\
\label{camera_area}
\end{split}
\end{equation}

To satisfy the optical objective of suppressing ambient-light, the camera must be placed outside the areas where these rays can arrive.
This makes the camera be put upper and more right, strongly restricting the compactness of the sensor. For example, when $\theta_{tv}=\theta_c$, it makes $\theta_{iv}=\theta_{ev}=0$ and $\theta_{cam}=f(\theta_{iv})= \frac{\pi}{2}+\theta_{c}$.
In addition, to avoid these rays being reflected into the camera again, special design, such as matte-black surfaces, is required for the container (mount) of the camera.
Therefore, to ensure more reliable satisfaction of the optical objective and more compact design, we finally choose $\theta_{tv}>2\theta_c$ for \sensor.

\subsection{Design of Sensor Components}
\label{app:sensor_components}
\begin{figure}[ht]
\centering
\includegraphics[width=0.9\linewidth]{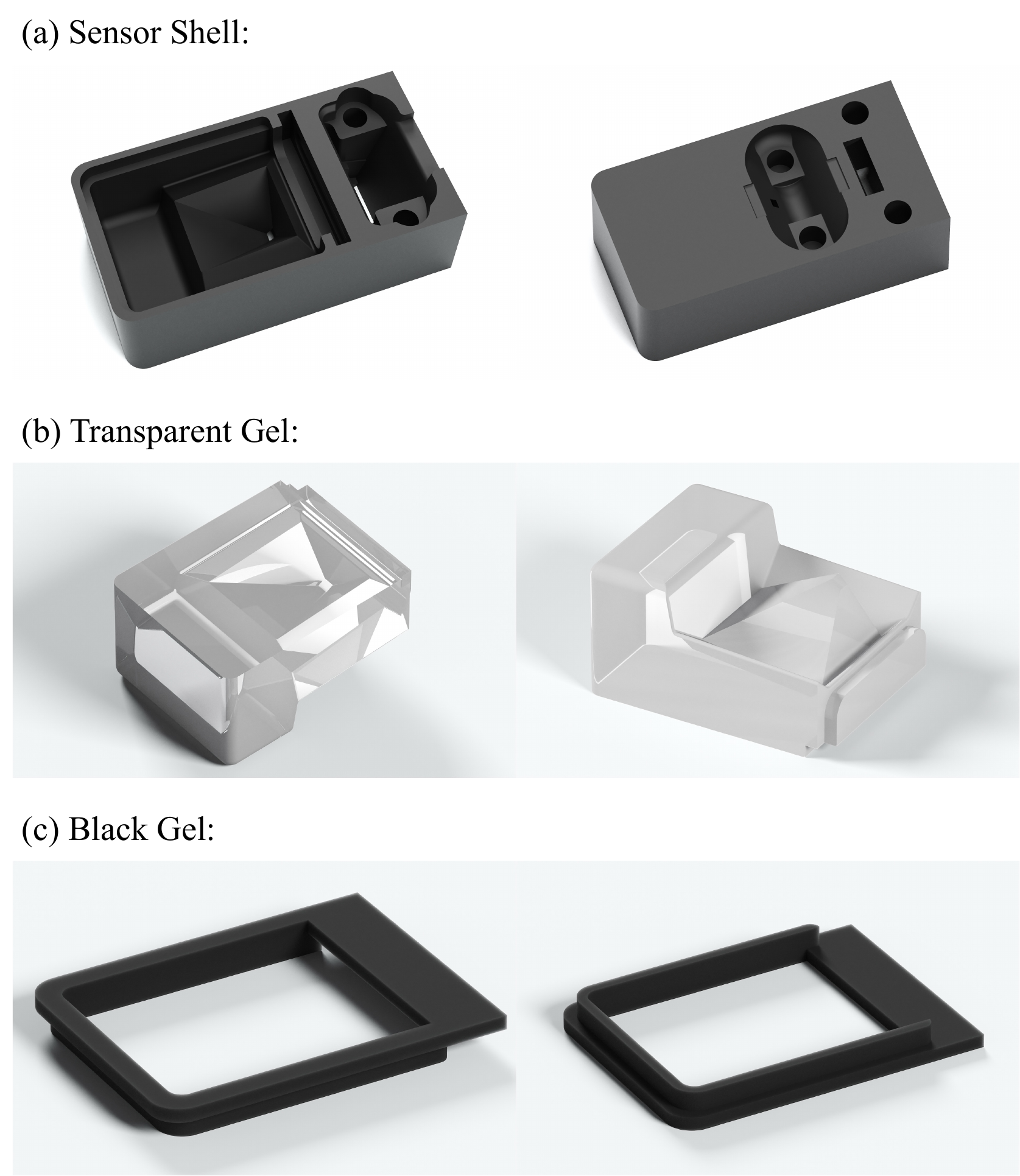}
\vspace{-0.2cm}
\caption{
Detailed views of the sensor shell, transparent gel and black gel.
}
\label{fig_app:design_detail}
\end{figure}

We provide more rendered views of the sensor shell, transparent gel and black gel in Fig.~\ref{fig_app:design_detail}.

\subsection{Optical Analysis with Wedge Shape ($\theta_s<\frac{\pi}{2}$)}
\label{app:wedge_shape}
For certain robotic applications, \sensor may benefit from a wedge-shaped profile to access cluttered or confined spaces.
To study the optical implications, we design and fabricate four wedge-shaped \sensor variants with different shell angles $\theta_s$, using the same fabrication procedure as the default design.
Under a bright indoor environment (approximately $1000~\text{Lux}$), we record non-contact images from each variant and report their grayscale mean and standard deviation in Table~\ref{tab:wedge_shape}.

\newcommand{\widthwedge}{1.8cm}
\newcommand{\rownumwedge}{4}
\begin{table}[ht]
\centering
\setlength{\tabcolsep}{2pt}
\caption{Wedge-shaped \sensor variants and the mean and standard deviation of their non-contact images}
\begin{tabular}{lcccc}
\toprule
$\theta_s$ ($^\circ$) & 45 & 40 & 35 & 30 
\\
\multicolumn{1}{l}{\multirow{\rownumwedge}{*}{Shell}}
& \multirow{\rownumwedge}{*}{\includegraphics[width = \widthwedge]{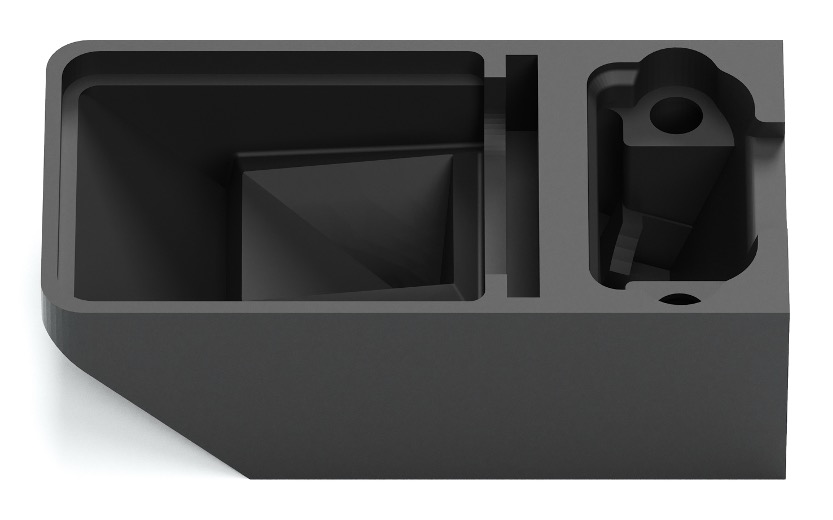}}
& \multirow{\rownumwedge}{*}{\includegraphics[width = \widthwedge]{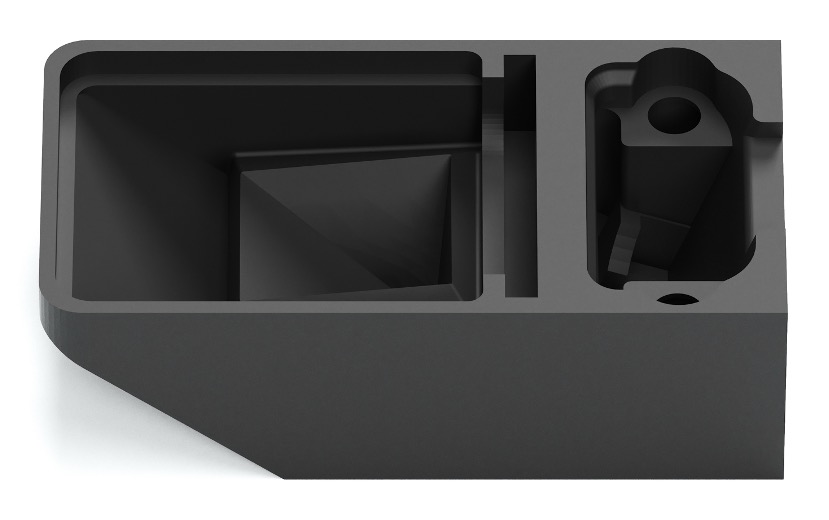}}
& \multirow{\rownumwedge}{*}{\includegraphics[width = \widthwedge]{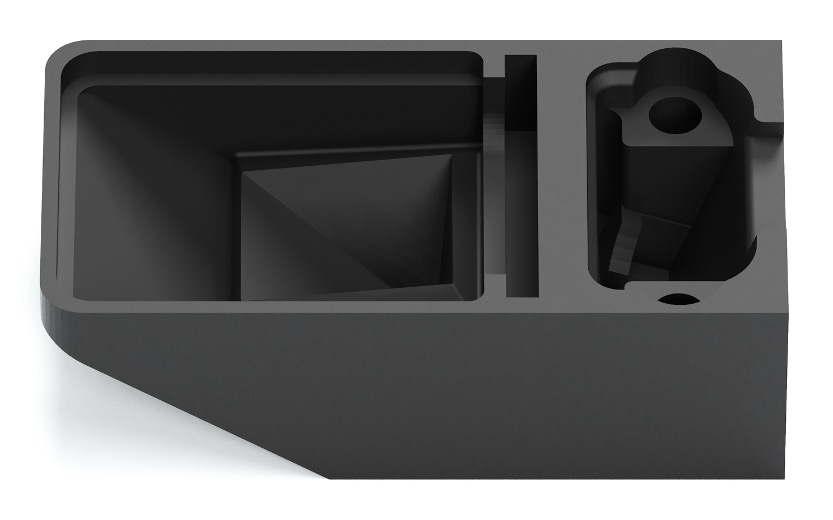}}
& \multirow{\rownumwedge}{*}{\includegraphics[width = \widthwedge]{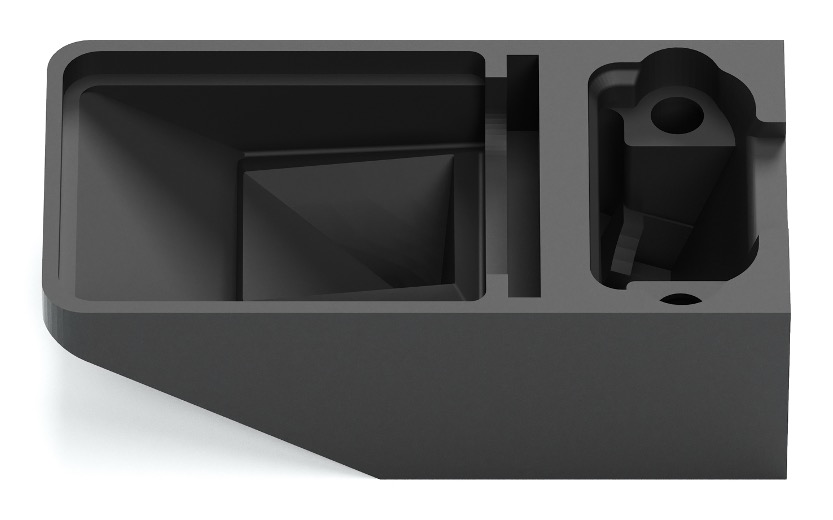}}
\\ \\ \\ \\
\midrule
Mean & 2.10 & 2.47 & 2.91 & 3.20
\\
Std & 0.85 & 1.03 & 1.26 & 2.33
\\
\bottomrule
\end{tabular}
\label{tab:wedge_shape}
\end{table}

As $\theta_s$ decreases, the slanted internal shell surface becomes more likely to reflect ambient light (both external and internal) into directions that can enter the camera through non-contact regions, leading to a gradual increase in non-contact intensity.
Nevertheless, even at $\theta_s=30^\circ$, the non-contact images remain close to black (mean gray value $<4.0$), indicating that the light-suppression property is largely preserved under moderate wedge shaping.
If even darker non-contact images are desired, the slanted surface can be further improved by applying higher-absorption black coating or by adding micro-textures/light-trap patterns to reduce specular reflections.

\begin{figure*}[t]
\centering
\includegraphics[width=0.8\linewidth]{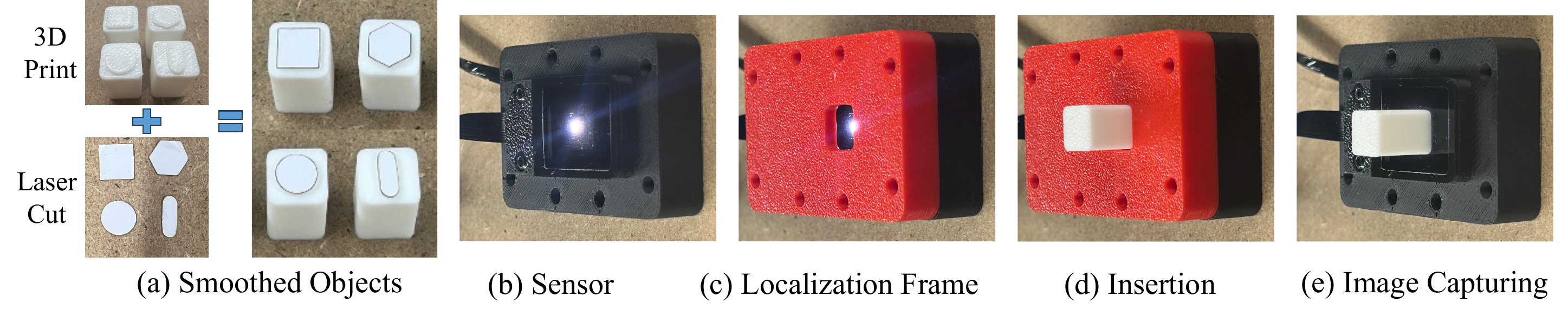}
\vspace{-0.2cm}
\caption{\small{Quantitative evaluation setup for contact segmentation. A localization frame constrains the pose of laser-cut planar shapes, enabling automatic generation of pixel-level ground-truth masks.}}
\label{fig:evaluation_process}
\end{figure*}

\begin{figure*}[t]
\centering
\includegraphics[width=0.8\linewidth]{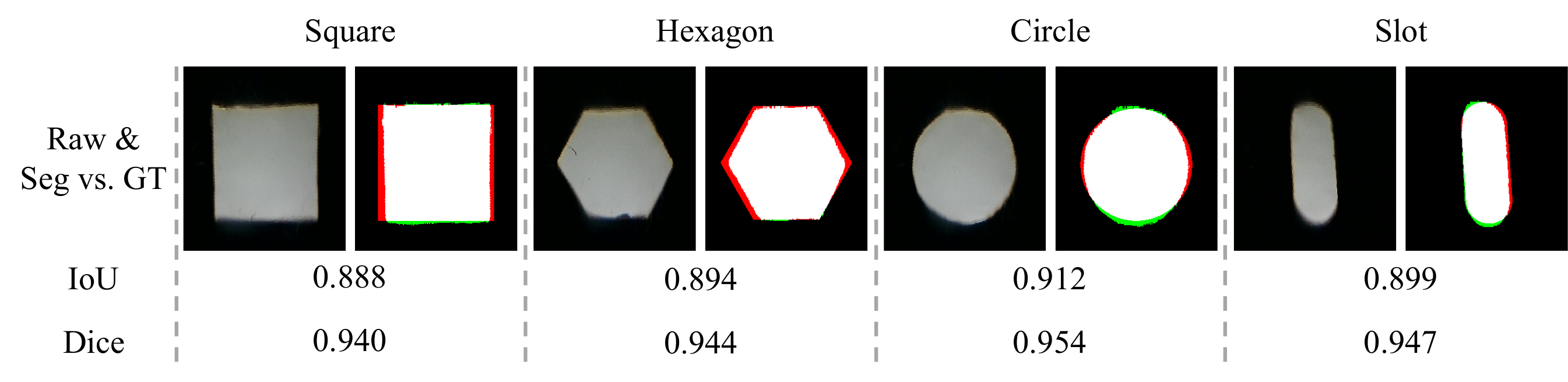}
\vspace{-0.2cm}
\caption{\small{Quantitative contact segmentation results on four geometries. For each shape, we show the raw tactile image and the overlay between segmentation and ground truth (\textcolor{green}{green}: segmentation; \textcolor{red}{red}: ground truth; white: overlap).}}
\label{fig:evaluation_results}
\end{figure*}

\begin{figure}[ht]
\centering
\includegraphics[width=0.8\linewidth]{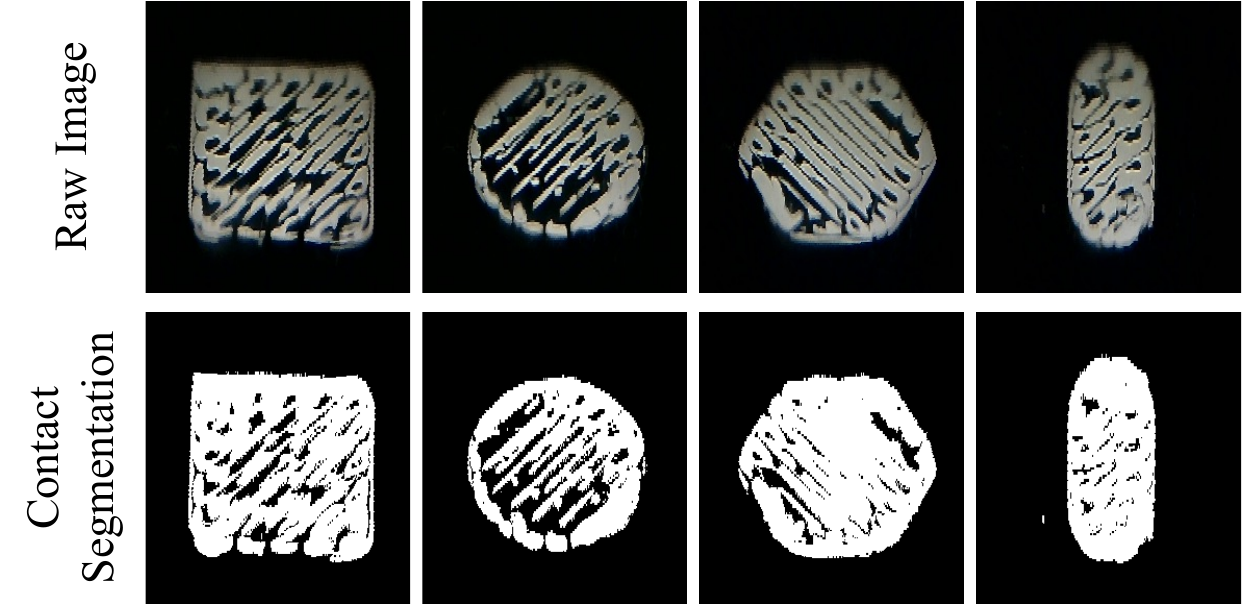}
\vspace{-0.2cm}
\caption{\small{Raw 3D-printed parts are unsuitable for exact ground-truth generation because \sensor captures their surface imperfections.}}
\label{fig:evaluation_print_only}
\end{figure}

\subsection{Quantitative contact segmentation.}
\label{quantitative_segmentation}
To quantitatively evaluate deformation-independent contact segmentation, we design a controlled protocol that enables the automatic generation of pixel-level ground-truth masks (Fig.~\ref{fig:evaluation_process}).
We use four standard planar geometries: square, hexagon, circle, and slot.
For each geometry, we laser-cut a smooth tape with the corresponding shape, attach it to a matched holder, and use a localization frame to constrain its contact pose relative to the sensor.
Since the tape geometry and contact pose are fixed by design, the corresponding ground-truth mask can be automatically generated from the designed shape, while the segmentation algorithm itself takes only the tactile image as input.

Across the four geometries, \sensor achieves IoU and Dice scores above $0.888$ and $0.940$, respectively (Fig.~\ref{fig:evaluation_results}).
Qualitatively, the predicted masks are well aligned with the raw tactile images; the remaining metric errors are mainly caused by small shape inaccuracies introduced during laser cutting.
These results provide quantitative evidence that \sensor can accurately segment contact regions without relying on visible deformation.

We use smooth tapes rather than raw 3D-printed parts because \sensor is sensitive enough to capture surface imperfections from printed surfaces, as shown in Fig.~\ref{fig:evaluation_print_only}.
These imperfections make the nominal CAD geometry of 3D-printed parts unsuitable as exact ground truth for pixel-level segmentation evaluation.

\subsection{Sensing Limitations of \sensor}
\label{app:sensing_limitations}
Although \sensor enables deformation-independent contact sensing under everyday lighting, its current optical design still has several limitations. We discuss them below to clarify the intended sensing regime and motivate future improvements.

\textbf{Extremely dark materials.}
\sensor detects contact by capturing contact-induced light scattered from the contacting surface. Therefore, the sensing signal depends partly on how much light the contacted material can scatter or reflect toward the camera. Extremely dark or highly light-absorbing materials may return very little light even when they are in contact with the sensor, reducing the contrast between contact and non-contact regions. In such cases, contact segmentation can become less reliable.

\textbf{Highly specular materials.}
Highly specular or mirror-like materials can also be challenging for \sensor. Instead of producing sufficient diffuse scattering toward the camera, incident light may be strongly reflected away from the imaging path or concentrated into saturated highlights. As a result, the captured image may contain weak, incomplete, or locally overexposed contact signals, which can degrade segmentation quality. Future designs may mitigate this limitation through improved illumination geometry, polarization-based filtering, or multi-view optical layouts.

\textbf{Sensitivity to dirt or residue.}
Because \sensor is designed to detect extremely light contact, it is also sensitive to small particles, residue, or dirt on either the sensing surface or the contacted object. Such contaminants can scatter light locally and may be mistakenly segmented as contact. In practice, this sensitivity is a trade-off of the proposed sensing principle: the same high sensitivity that enables near-zero-force contact detection also makes the sensor responsive to unintended surface contamination. Therefore, keeping the sensing surface clean is important for reliable operation, especially in tasks that require precise contact masks.

\textbf{Force and shear estimation.}
The current implementation of \sensor focuses on contact detection and contact segmentation, rather than explicit normal-force or shear-force estimation. In principle, marker-based force sensing could be added because \sensor uses a transparent deformable gel. However, printed markers would provide sparse deformation measurements and partially occlude the appearance-preserving contact image. More importantly, deformation-based force inference becomes weak in the extremely light-contact regime targeted by \sensor, where visible deformation can be negligible or absent. A promising future direction is to develop methods that infer richer force-related information while preserving the current deformation-independent contact sensing capability.

\subsection{Ablation Study For Sensor Design}
\label{app:design_ablation}

\begin{figure}[ht]
\centering
\includegraphics[width=0.56\linewidth]{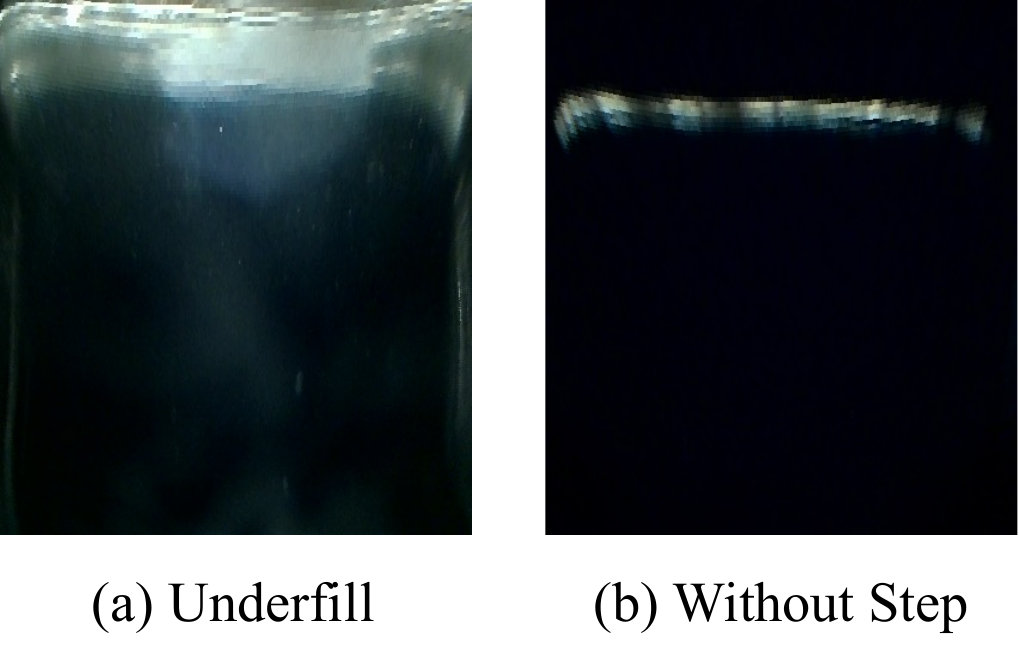}
\vspace{-0.2cm}
\caption{
Underfill and the missing perimeter step of the black gel both allow external light to leak into the sensor.
}
\label{fig_app:abla_side}
\end{figure}

\paragraph{Acrylic mold for casting the black gel}
Without the acrylic mold, the narrow perimeter cavity makes it difficult to control the volume of black silicone, often resulting in either overflow or underfill.
Overflow contaminates the sensing surface and permanently blocks contact visibility in the covered regions, while underfill exposes side surfaces and opens a direct leakage path for ambient light to reach the camera (Fig.~\ref{fig_app:abla_side} (a)).

\paragraph{Edge indentation on the shell outer walls}
As shown in Fig.~\ref{fig_app:abla_side} (b), removing the perimeter step (via the edge indentation design) prevents the black gel from forming a light-blocking lip at the shell boundary.
This leaves a thin gap between the shell and the gel, which again creates a leakage path for ambient light into the sensor and raises the non-contact background.

\begin{figure}[ht]
\centering
\includegraphics[width=0.48\linewidth]{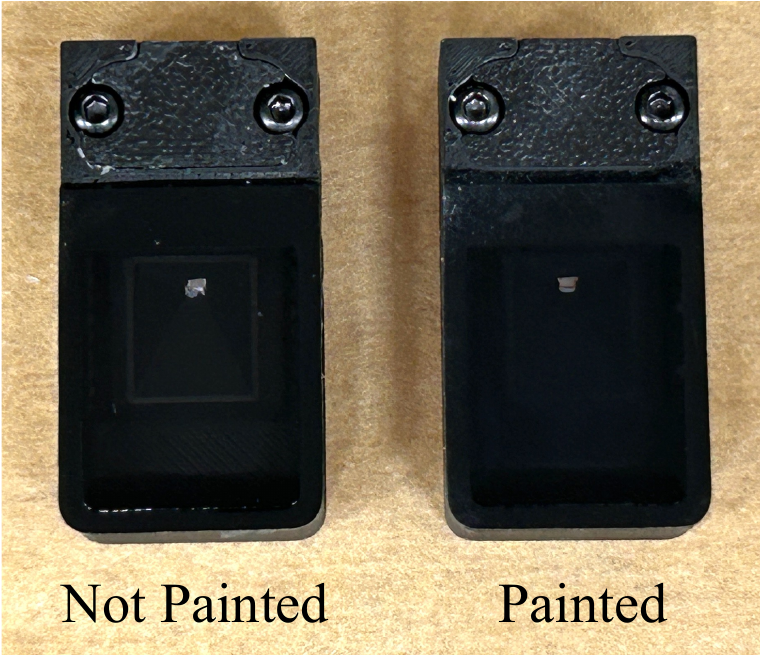}
\vspace{-0.2cm}
\caption{
Applying high-absorption black paint substantially darkens the shell inner walls and reduces stray reflections.
}
\label{fig_app:abla_paint}
\end{figure}

\paragraph{Black paint on the shell inner walls}
As shown in Fig.~\ref{fig_app:abla_paint}, brushing high-absorption black paint on the inner walls significantly darkens the internal surfaces.
In practice, the unpainted shell reflects more stray light toward the camera as ambient brightness increases, leading to a brighter non-contact background and reduced contrast.

\subsection{Large-Area Contact Sensing}
\label{app:large_area}
\begin{figure}[ht]
\centering
\includegraphics[width=\linewidth]{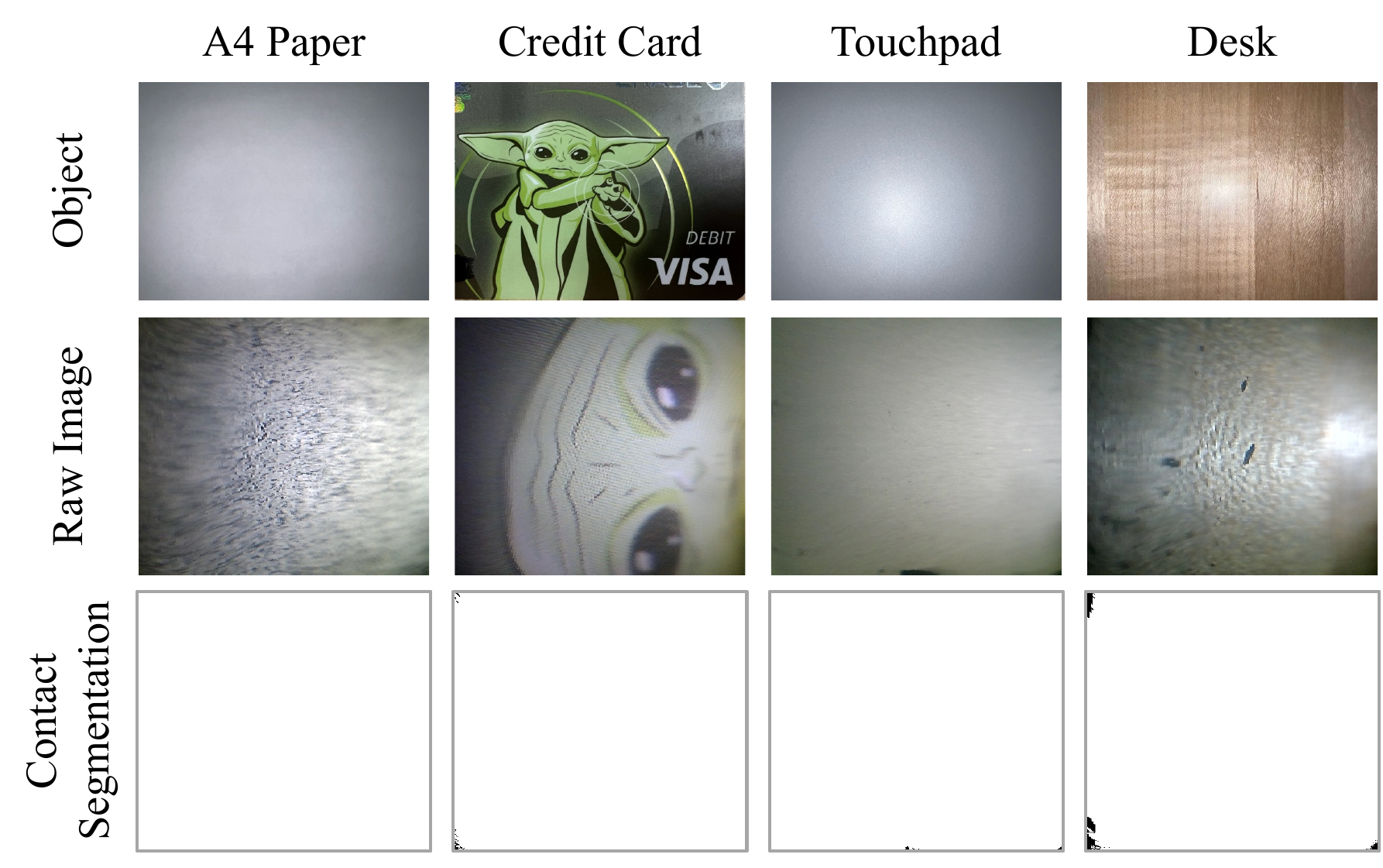}
\vspace{-0.6cm}
\caption{
\sensor sensing large-area contact on flat surfaces.
Unlike deformation-based VBTSs, \sensor does not rely on measurable indentation and therefore remains effective even when contact is broad and deformation is minimal.
}
\label{fig_app:large_area}
\end{figure}

Large-area contact, especially against flat surfaces, is often challenging for deformation-based VBTSs because the resulting deformation is both slight and spatially uniform, yielding weak or ambiguous signals.
As shown in Fig.~\ref{fig_app:large_area}, \sensor avoids this limitation since its deformation-independent, optics-based principle makes contact directly visible, allowing it to reliably detect and segment large-area contact on flat surfaces without requiring heavy pressing.

\subsection{Baseline Vision-Based Tactile Sensors}
\label{app:baseline_sensor}
\begin{figure}[ht]
\centering
\includegraphics[width=\linewidth]{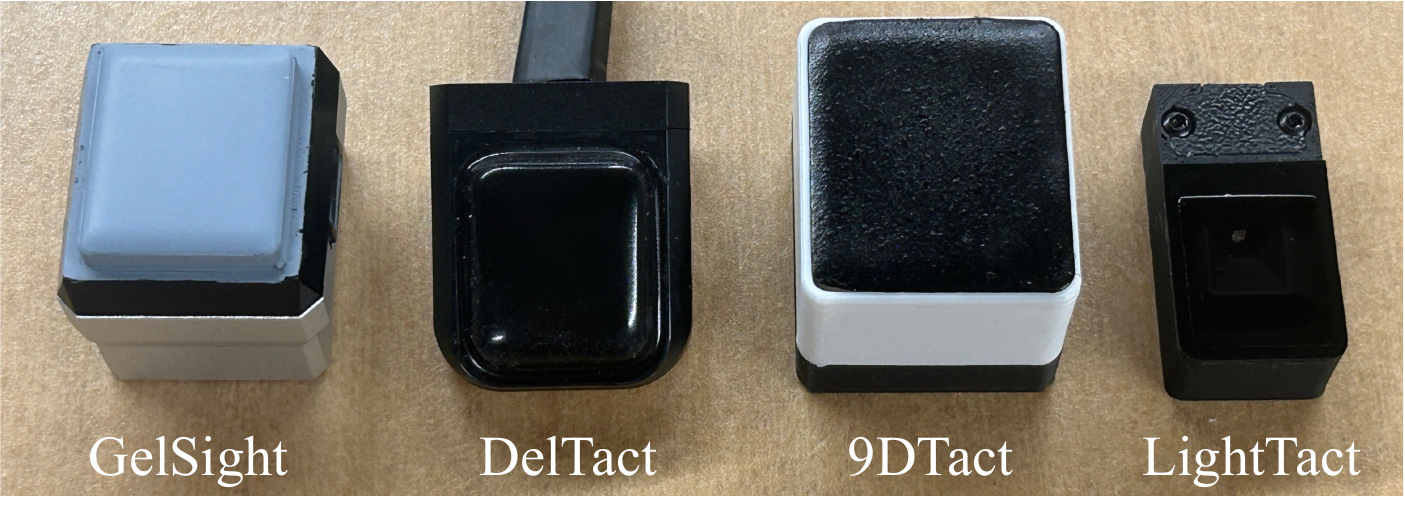}
\vspace{-0.6cm}
\caption{
Three baseline sensors and \sensor.
}
\label{fig_app:baseline_sensors}
\end{figure}

Our baseline VBTSs include GelSight, DelTact and 9DTact, as shown in Fig.~\ref{fig_app:baseline_sensors}.

\subsection{Details for Resistor Sorting}
In the resistor sorting task, our text prompt to VLMs is:
\begin{mdframed}
The attached images are captured by two tactile sensors, which are mounted on the left and right side of a gripper.
They are grabbing a resistor together.
The tactile sensors can only see the part of the resistor that is contacting with the sensor's sensing surface.
For the rest part of the sensing surface that is no in contact, the pixels remain black.
The resistor has 5 bands that represent its resistance value. Please carefully analyze the 5 colors of the bands and compute the final resistance value of the resistor accurately.
Notice that, read the color bands of the left sensor's image from left to right. Whereas read the color bands of the right sensor's image from right to left.
Note that the internal illumination of the tactile sensor is relatively bright so be careful when inferring the colors.
Please only output the final resistance value.
\end{mdframed}

\end{document}